\documentclass[10pt,journal,compsoc]{IEEEtran}
\usepackage{natbib}
\setcitestyle{square,sort,comma,numbers}
\usepackage{multirow, booktabs}
\usepackage{siunitx}
\usepackage{makecell}

\usepackage{mathrsfs}

\usepackage{graphicx}

\usepackage[font=small,labelfont=bf]{caption}
\usepackage{amsthm}
\usepackage{pifont}

\usepackage{algorithmic}
\usepackage{algorithm}

\usepackage{amsmath}
\usepackage{epstopdf} 
\usepackage{subcaption}
\usepackage{lipsum}
\usepackage{amssymb,amsmath,amsthm}

\newtheorem{lemma}{Lemma}

\newtheorem{problem}{Problem}
\newtheorem{definition}{Definition}

\usepackage{mathtools} % Bonus

\usepackage{hyperref}
\hypersetup{
	colorlinks=true,
	linkcolor=blue,
	filecolor=blue,      
	urlcolor=blue,
	citecolor=blue
}

\begin{document}
	
	\newcommand*{\everymodeprime}{\ensuremath{\prime}}
	\title{EmoDNN: Understanding emotions from short texts through a deep neural network ensemble}
	
	\author{Sara~Kamran,
		Raziyeh~Zall,
		Mohammad~Reza~Kangavari,
		Saeid~Hosseini,
		Sana~Rahmani, and
		Wen~Hua
		%~\IEEEmembership{Life~Fellow,~IEEE}% <-this % stops a space

		\IEEEcompsocitemizethanks{
			\IEEEcompsocthanksitem S. Kamran, R. Zall, M. R. Kangavari, and S. Rahmani are with School of computer engineering, Iran University of Science and Technology. \protect\\Email: {sara.kamran72}@gmail.com, {zall\_razieh}@comp.iust.ac.ir, {kanagvari}@iust.ac.ir, {rahmany.sana}@gmail.com
			\IEEEcompsocthanksitem S. Hosseini is with the Faculty of Computing and Information Technology, Sohar University, Oman. Email: {sahosseini}@su.edu.om
			\IEEEcompsocthanksitem W. Hua is with the School of Information Technology and Electrical Engineering, University of Queensland, Australia. E-mail: {w.hua}@uq.edu.au
		}
		\thanks{}}
	
	
	\markboth{}%
	{Shell \MakeLowercase{\textit{et al.}}: Bare Demo of IEEEtran.cls for Computer Society Journals}
	
	
	
	\IEEEtitleabstractindextext{%
		\begin{abstract}			
The latent knowledge in the emotions and the opinions of the individuals that are manifested via social networks are crucial to numerous applications including social management, dynamical processes, and public security. Affective computing, as an interdisciplinary research field, linking artificial intelligence to cognitive inference, is capable to exploit emotion-oriented knowledge from brief contents. The textual contents convey hidden information such as personality and cognition about corresponding authors that can determine both correlations and variations between users. Emotion recognition from brief contents should embrace the contrast between authors where the differences in personality and cognition can be traced within emotional expressions. To tackle this challenge, we devise a framework that, on the one hand, infers latent individual aspects, from brief contents and, on the other hand, presents a novel ensemble classifier equipped with dynamic dropout convnets to extract emotions from textual context. To categorize short text contents, our proposed method conjointly leverages cognitive factors and exploits hidden information. We utilize the outcome vectors in a novel embedding model to foster emotion-pertinent features that are collectively assembled by lexicon inductions. Experimental results show that compared to other competitors, our proposed model can achieve a higher performance in recognizing emotion from noisy contents.
		\end{abstract}
		\vspace{-3mm}
		}		
		\maketitle
		\IEEEpeerreviewmaketitle
		\IEEEraisesectionheading{\section{Introduction}\label{introductionSection}}
		\IEEEPARstart{U}{nderstanding} emotional information from short text content finds important applications in numerous domains: (i) In conversation transcripts in user contents \cite{Phan2018}. (ii) In the political context, to foster the prediction of the ballet results \cite{Budiharto2018}. (iii) In health, to better recognize the people affected by extreme depressions \cite{Corazza2020}\cite{Desmet2013} (iv) In sales and finance, to enhance the product development \cite{Ullah2016} and predict the market fluctuations \cite{Bollen2011}. Nowadays, with the spread of social networks, people share their brief contents conveying latent cues such as personality that can identify the contrast between authors. Given a single short-text $s_i$ with corresponding cognitive cues $p_j$, we aim to identify the emotions explaining $s_i$. Compared to comprehensive formal documents, individuals usually tend more to reveal their instant emotions through composing informal social media posts. Accordingly, we can utilize such rich content to exploit real-world emotion-related social communities. However, challenges abound:\\
		\noindent\textit{\textbf{Challenge 1 (Ignoring Author Latent Information)}}\\
		\indent Numerous prior works \cite{Esmin2012}\cite{Jain2017}\cite{Xu2015}\cite{Colneric2018} identify the emotions from textual contents disregarding the individual differences between authors, while such contrast including personality can affect the way people express their emotions \cite{Watson1992}. As Holtgraves et al. \cite{Holtgraves2011} report, the personality-based correlations are more likely manifested in the emotion-pertinent expressions. Table \ref{tab:AuthorRelevance} demonstrates the correlation between a tweet and relevant emotion and cognitive vectors. Here emotion vector comprises the score for each emotion [anger, disgust, etc.] where we use a similar vector to express personality through Extraversion, Openness, etc. The respective binary and floating values in the emotion and personality vectors imply whether the short-text corresponds to each emotion or cognitive factors.\\
		\begin{table*}
			\centering
			\def\arraystretch{1.5}
			\tiny
			\vspace{-2mm}
			\begin{tabular}{|l|l|l|l|l|l|l|l|l|l|l|}
				\hline
				& \multicolumn{5}{c|}{Emotion}                                                                                           & \multicolumn{5}{c|}{Cognitive Factors}                                                                                        \\ \hline
				Tweet                                           & anger                 & disgust               & fear                  & joy                   & sadness               & OPE.                  & CON.                  & EXT.                  & AGR.                  & NEU.                  \\ \hline
				I've never been so excited to start a semester! &  &  &  & \checkmark &  & 4.1 & 3.5 & 3.38 & 3.623 & 2.559 \\ \hline
				I'm a shy   person                              &  &  &  &  & \checkmark & 4.11 & 3.498 & 3.392 & 3.622 & 2.561 \\ \hline
				I am just so bitter today  & \checkmark & \checkmark &  &  & \checkmark & 4.0 & 3.508 & 3.366 & 3.564 & 2.58 \\ \hline
			\end{tabular}
			\vspace{-2mm}
			\caption{Conceptual relevance}
			\label{tab:AuthorRelevance}
			\vspace{-7mm}
		\end{table*}
		\begin{figure}[t]
			\begin{subfigure}{.47\linewidth}
				\centering
				% include first image
				\includegraphics[width=1\linewidth]{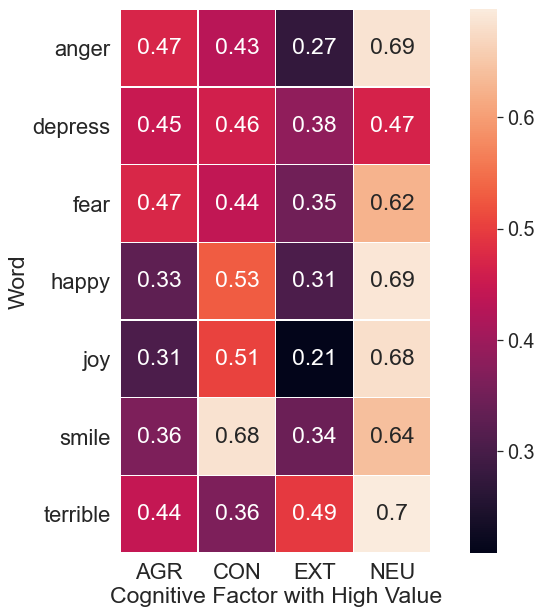}  
				\vspace{-5mm}
				\caption{High Cognitive Factors}
				\label{fig:High_Cog_word}
			\end{subfigure}
			\begin{subfigure}{.47\linewidth}
				\centering
				% include second image
				\includegraphics[width=1\linewidth]{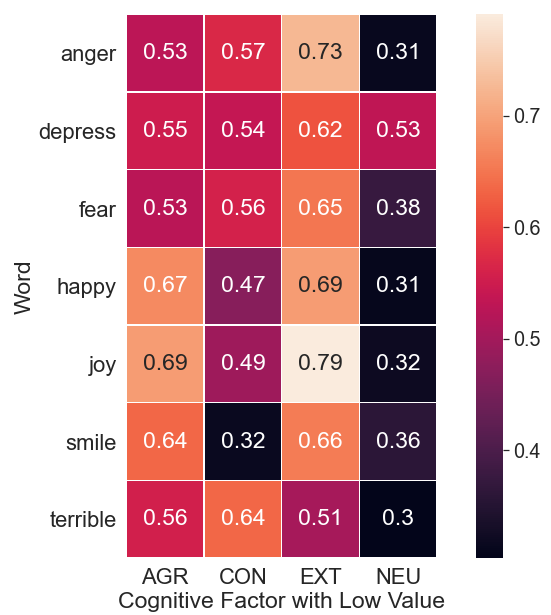}  
				\vspace{-5mm}
				\caption{Low Cognitive Factors}
				\label{fig:Low_Cog_word}
			\end{subfigure}
			\vspace{-2mm}
			\caption{Emotion words connect to cognitive factors}
			\label{fig:word_cog_emo}
			\vspace{-8mm}
		\end{figure}
		\vspace{-5mm}
		\par\noindent
		To investigate how human personality can affect the emotion manifest in expressions, we consume the SemEval dataset \cite{Mohammad2018} to set up an observation based on the distribution probabilities of the words. We select the expressions with various emotional intensities that are further pertinent to diverse cognitive factors. As Fig. \ref{fig:word_cog_emo} demonstrates, the distribution probabilities for each emotional word differ in various cognitive factors. Users with high NEU (Neuroticism) have low emotional stability and frequently use distressing words, like terrible. So their short text contents include more emotional words compared to the user of the low NEU weights. Interestingly, users with low EXT (Extraversion) are reluctant to interact with others and tend to express negative emotions such as fear, depression, and anger more frequently than other users with high EXT load. Individuals with a high rate of CON (Conscientiousness) use a greater number of positive emotional words, such as happy, joy, and smile in their contents. From another perspective, people with high AGR (Agreeableness) weights rarely utilize emotional-related words. Therefore, the proposed consensus results reveal how cognitive factors can influence emotional expressions in short-text contents. Since individuals with identical cognitive factors exhibit similar emotions, we leverage the latent aspects of personalities to enhance our ability to detect emotions.\\
		\noindent\textit{\textbf{Challenge 2 (Noisy Short Text)}}\\
		\indent Short texts include important information but they are brief and error-prone. Hence, it is a tedious task to associate emotion cues with cognitive features.\\
		\noindent\textit{\textbf{Challenge 3 (The Scarceness of Annotated Dataset)}} \\
		\indent All prevailing datasets have either emotional labels \cite{Mohammad2017}\cite{Mohammad2018} or cognitive labels \cite{Kosinski2013}. So, due to the scarceness of the dataset which contains both emotion and cognitive annotations, one of our challenges is to create datasets that are annotated by emotion and cognitive cues.\\
		\noindent\textbf{Contributions.} While our prior works \cite{Najafipour2020}\cite{Hosseini2020} handle short text contents to detect attributed segments \cite{Hosseini2014} and identify concepts, removing the perturbation caused by external knowledge bases, this paper recognizes the emotions through leveraging the cognitive factors from similar brief context. To this end, we utilize cognitive factors of individuals to categorize short texts to feed our novel ensemble classifier. Furthermore, in our proposed framework, we detect the emotion embedding of short texts with the help of an external knowledge-base which is associated with emotion lexicons. Accordingly, we extract multi-channel features in short texts to enrich the short-text inference models. Our contributions are fourfold:
		\vspace{-2mm}
		\begin{itemize}			
			\item We develop a framework on emotion recognition through ensemble learning which leverages the cognitive cues to distinguish between short-text authors.			
			\item We propose a regression approach for inferring latent cues about short-text authors.
			\item We design a multi-channel feature extraction algorithm based on emotion lexicons and attention mechanisms that fuse various embedding models to retrieve preferable vectors.
			\item We devise a cognitive aware aggregation function to combine the results of different base classifiers in the ensemble model. 
		\end{itemize}
		\vspace{-2mm}
		\par\noindent
		The rest of our paper is as follows: in Sec. \ref{relatedwork}, we study the literature; in Sec \ref{problemstatement}, we provide the problem and our framework; in Sec. \ref{methodology} and \ref{experiments}, we respectively explain our model and experiments. The paper is concluded in Sec. \ref{conclusion}.
		\vspace{-8mm}
		\section{Related Work}
		\label{relatedwork}
		\vspace{-2mm}
		As briefed in Table \ref{relatedwork}, the related work comprises personality prediction and emotion recognition.
		\vspace{-5mm}
		\subsection{Personality Prediction}
		\label{Sec_Word_Embedding_related}		
		\vspace{-1mm}
		Personality is a psychological construct and it aims to understand various human behaviors with stable and measurable individual characteristics \cite{Xue2018}. Various studies predict and identify personality traits \cite{Kosinski2013} from social networks. The methods for cognitive prediction are of two categories: 1) Machine learning methods; 2) Deep learning models.\\ General machine learning algorithms \cite{Golbeck2011} leverage various features \cite{Hosseini2018} including linguistic stats such as word count \cite{Alam2013} and social network attributes like the number of friends to estimate personality traits \cite{Quercia2011}.	Deep Neural network approaches \cite{Yuan2018}\cite{Sun2018}\cite{Xue2018}\cite{Majumder2017} surpass traditional methods. Sun et al. \cite{Sun2018} combine bidirectional LSTMs (Long Short Term Memory networks) with CNNs (Convolutional Neural Network) to infer structures of texts. Similarly, \cite{Xue2018} devise AttRCNN structure to understand the semantic features that are amended by the statistical linguistic features. However, \cite{Majumder2017} utilizes essential statistics including Mairesse features, the number of words, and an average length of sentences that are combined within convnets in a hybrid manner. We initially exploit latent personal characteristics by a modified Support Vector Regression (SVR) model and subsequently adopt CNN convents to extract feeling sentiments.\\
		\vspace{-4mm}
		\begin{table}[H]
			\label{tab:literature}
			\centering
			\vspace{-2mm}
			\caption{Literature}
			\vspace{-3mm}
			\def\arraystretch{1.5}
			\tiny
			\begin{tabular}{lll}
				\Xhline{2\arrayrulewidth}
				Category                                                                          & Approaches                        & references              \\ \hline
				\multirow{2}{*}{\begin{tabular}[c]{@{}l@{}}Personality\\ Prediction\end{tabular}} & Machine Learning                  & \cite{Golbeck2011}\cite{Alam2013}\cite{Quercia2011} \\ \cline{2-3} 
				& Deep Learning                     & \cite{Xue2018}\cite{Yuan2018}\cite{Sun2018}\cite{Majumder2017}        \\ \hline
				\multirow{5}{*}{\begin{tabular}[c]{@{}l@{}}Emotion\\ Recognition\end{tabular}}    & Lexicon-Based                     & \cite{Mohammad2017a}\cite{Araque2019} \\ \cline{2-3} 
				& \multirow{4}{*}{Machine Learning} & \cite{Esmin2012}\cite{Jain2017}\cite{Xu2015}\cite{Li2014} \\
				&  & \cite{Bandhakavi2017}\cite{Udochukwu2015}\cite{ghazi2014prior} \\
				&  & \cite{Canales2017}\cite{Halim2020} \\ \cline{2-3} 
				& \multirow{2}{*}{\begin{tabular}[c]{@{}l@{}}Deep Learning\end{tabular}}                     & \cite{Phan2018}\cite{Colneric2018}\cite{Deng2020}\cite{Batbaatar2019} \\ 
				& &\cite{Akhtar2019}\cite{Cai2019}\cite{Wang2020}\cite{Rong2019}
				\\ \Xhline{2\arrayrulewidth}\\
			\end{tabular}
			\vspace{-7mm}
		\end{table}				
		\vspace{-4mm}		
		\subsection{Emotion Recognition}
		\vspace{-1mm}
		Affective Computing \cite{Picard2000} as an emerged field of research has attracted significant attention. Because it results in systems that can automatically	recognize human emotions and eminently influence decision-making procedures. Emotions can be detected by heuristic, machine learning, and deep learning approaches \cite{Deng2021}. The lexicon-based heuristic approaches \cite{Mohammad2017a}\cite{Araque2019} find keywords in textual contents and assign emotion labels based on lexicon tags, referenced from knowledge-base tools such as NRC-EIL \cite{Mohammad2017a} and DepecheMood \cite{Araque2019}. Machine learning models \cite{Jain2017}\cite{Canales2017} not only consider lexicons but also extract effective textual features from input corpus, concluded by decision rules to recognize emotions based on the trained explicit labels. The ML models include Na\"ive Bayes (NB) \cite{Jain2017}, Random Forest (RF) \cite{Halim2020}, Support Vector Machine (SVM) \cite{Canales2017}\cite{Jain2017}\cite{Li2014}\cite{Halim2020}, Logistic Regression (LR) \cite{ghazi2014prior}, and some trending deep learning methods \cite{Colneric2018}\cite{Wang2020}.\\
		The Na\"ive Bayes classifiers \cite{Jain2017} efficiently utilize various word classes to perform prediction but result in lower accuracies. Random forests \cite{Halim2020} fuse a combination of tree-based predictors where each tree depends on the values of a random vector, sampled independently and with the same distribution from the trees in the forest. Random forest approaches are more efficient than the SVM methods \cite{Jain2017}\cite{Li2014} but empirically gain less accuracy. Both Esmin \cite{Esmin2012} and Xu \cite{Xu2015} et al. use hierarchical classification techniques to perceive emotion cues. Such techniques integrate three levels: neutrality versus emotionality, sentiment analysis, and emotion recognition. Utilizing domain-specific emotion lexicons, a combination of n-grams, and part of speech (pos) tagging features can foster the classification performance of the SVM modules \cite{Bandhakavi2017}. On the contrary, the intelligible rule-based methods share the goal of finding regularities in data, expressing the form of If-Else rules \cite{Udochukwu2015}. In emotion detection, the rule-based models infer emotion-related events that undertake the cause \cite{Udochukwu2015}.\\
		%Such methods provide better performances in causal emotion correspondences including product design and political evaluation.\\
		Deep neural network models \cite{Deng2020}\cite{Wang2020}\cite{Batbaatar2019}\cite{Phan2018} have promoted the performance of prior Natural Language Processing (NLP) techniques, including emotion analysis and recognition. CNN (Convolutional NN) \cite{Colneric2018}\cite{Batbaatar2019} and RNN (Recurrent NN)\cite{Colneric2018}\cite{Cai2019} are two common deep learning architectures that are often integrated at the top of embedding modules, e.g., GloVe and Word2Vec, to infer emotion-pertinent cues in textual contents. While the convnets can effectively extract n-gram features, they are not as productive as the RNN schemes in attaining correlation within long-term sequences. Nonetheless, CNN models are distorted toward subsequent context and neglect previous words. To address the issue, LSTM models \cite{Wang2020} exploit the intensity of emotions out of brief contents in a bidirectional manner which results in preferable outputs in single and multi-label classification tasks. The Deep Rolling model \cite{Rong2019} combines LSTM and CNN into an ensemble to create a non-linear emotion-prediction model. Absorbed by the appealing performance of the deep neural network models \cite{Deng2021}, we devise a novel ensemble classifier equipped with dynamic dropout convnets that further leverages individual latent aspects, known as cognitive cues. Moreover, we propose a nontrivial method to extract features from emotion and semantic contents to feed the convnets in ensembles.\\		
		\vspace{-9mm}
		\section{Problem Statement}
		\label{problemstatement}		
		\vspace{-2mm}
		In this section, we elucidate preliminary concepts, problem statement for author linking, and our proposed framework.
		\vspace{-8mm}
		\subsection{Preliminary Concepts}
		\vspace{-2mm}
		\begin{definition} (short-text message)
			\label{def:shorttext_message}			
			$s_i \in S$ refers to a short-text that is composed by an author. Accordingly, $S$ is our corpus which includes all short-texts.
		\end{definition}	
		\vspace{-4mm}
		\begin{definition} (cognitive factors)  
			$p_i \in P$ ($P=\{p_i|i\in [1, ..., q]\}$) refers to a cognitive factor (e.g., neuroticism). Each factor reveals one latent perception for the given message.\\
		\end{definition}	
		\vspace{-8mm}
		\begin{definition} (emotion)  
			$e_i \in E$ ($E=\{e_i|i\in [1, ..., k]\}$) refers to a basic emotion such as happiness. Each emotion can be correlated with multiple cognitive factors.\\
		\end{definition}	
		\vspace{-8mm}
		\begin{definition} (cognitive category)
			Each cognitive category $c_i\in C$ can represent a set of short texts that are highly correlated based on pertinent cognitive factors, representing a hidden cognitive coupling.\\
		\end{definition}
		\vspace{-8mm}
		\begin{definition} (ensemble)
			$h_{c_i} \in H$ refers to a base classifier that is induced from cognitive category $c_i$($c_i\in C$). Accordingly, $H$ ($H=\{h_{c_1}, h_{c_2}, ..., h_{c_n}\}$) includes all base classifiers.
		\end{definition}
		\vspace{-4mm}
		\begin{definition} (emotion Vector)
			$\vec{V}_{s_i}=[v_{i,j}|j\in[1,...,k]]$ denotes the emotion scores for short text $s_i\in S$ where $v_{i,j}\in \mathbb{R}$ is an emotion score for $j^{th}$ basic emotion $e_j\in E$.
		\end{definition}
		\vspace{-4mm}
		\begin{definition} (cognitive Vector)
			$\vec{F}_{s_i}=[f_{i,j}|j\in[1,...,q]]$ denotes the cognitive factor scores for the short text $s_i\in S$ where $f_{i,j}$ is a cognitive score for $j^{th}$ factor $p_j\in P$. $\vec{F}_{s_i}$ represents short text $s_i$ in cognitive space with $q$ dimension according to $P$.
		\end{definition}
		\vspace{-7mm}
		\subsection{Problem Definition}
		\vspace{-1mm}
		\begin{problem} (identifying cognitive factors)
			Given a message $s_i$ our goal is to retrieve the cognitive vector $\vec{F}_{s_i}$ for $s_i$.\\
			\label{Identifying-Cognitive-Factors}
			\vspace{-2mm}
		\end{problem}		
		\vspace{-6mm}
		\begin{problem} (extracting multi-channel features)
			Given the set of short texts pertinent to a cognitive category $c_i$, our goal is to extract the set of features through tracking the emotional cues.\\
			\label{Extracting-multi-channel-features}
			\vspace{-2mm}
		\end{problem}		
		\vspace{-6mm}
		\begin{problem} (recognizing emotion through cognitive factors)
			Given the message $s_i$ and its associated cognitive vector $\vec{F}_{s_i}$, our goal is to construct the emotion vector $\vec{V}_{s_i}$.\\
			\label{Recognizing-Emotion-Through-Cognitive-Factors}
			\vspace{-2mm}
		\end{problem}
		\vspace{-8mm}
		\begin{figure}[htp]
			\centering
			\includegraphics[width=0.9\linewidth]{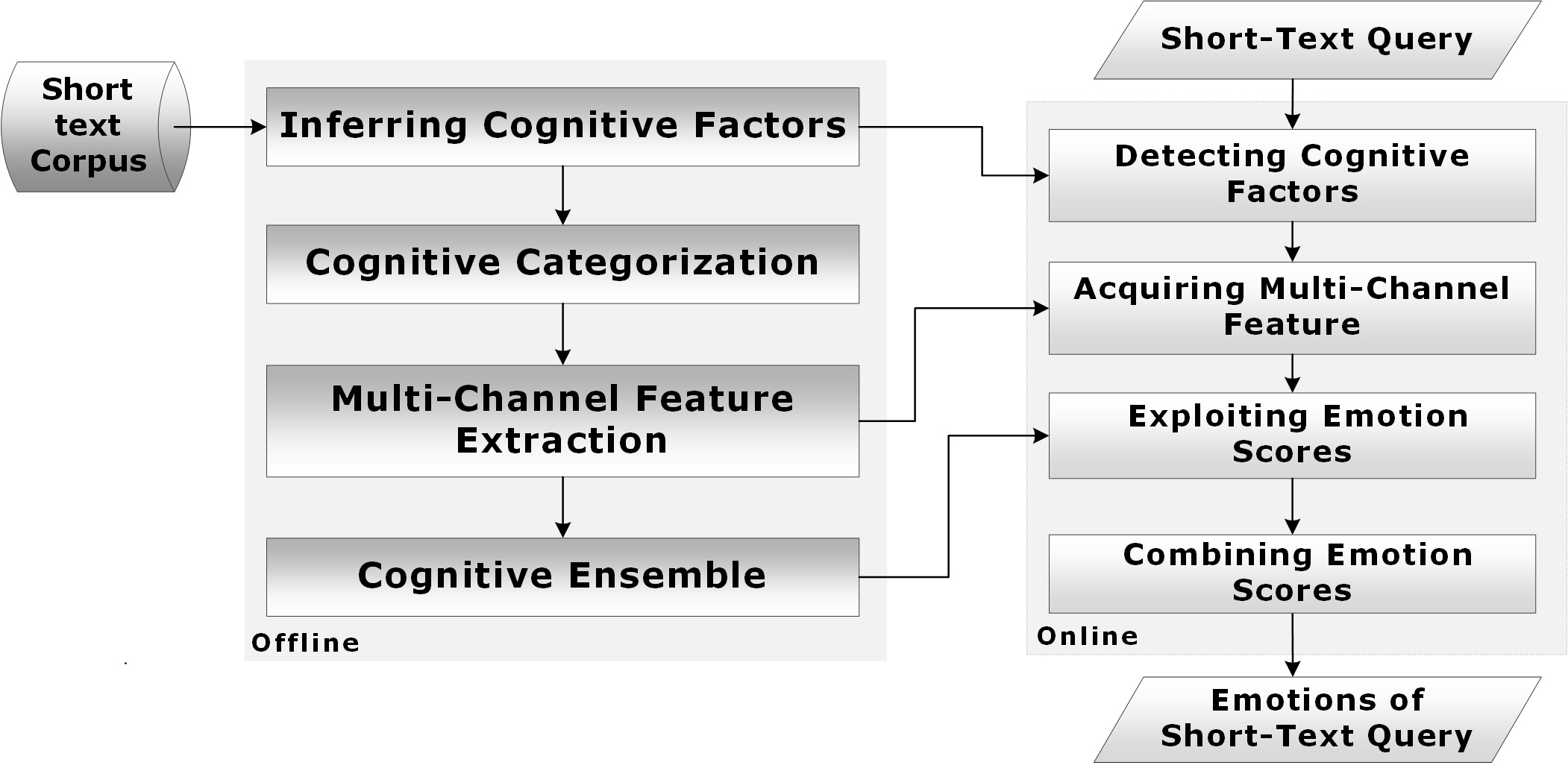}
			\vspace{-2mm}
			\caption{Framework}
			\label{fig:Framework}
			\vspace{-9mm}
		\end{figure}
		\vspace{-1mm}
		\subsection{Framework Overview}
		\label{Framework_Overview}
		\vspace{-1mm}
		The problem of emotion recognition through cognitive cues in brief contents includes two steps: (1) to extract the categories of short text messages via inclusive cognitive cues. (2) to learn an effective ensemble classifier to identify emotions by leveraging the extracted categories. Fig. \ref{fig:Framework} illustrates our proposed framework.\\
		In the \textit{offline} part, since a dataset enclosed with both emotion and cognitive annotations is scarce, as a prerequisite to emotion recognition, we initially augment the cognitive annotations within the dataset. To this end, we adopt the SVR to retrieve the cognitive vectors. Considering the impact of each cognitive factor on emotion-related expressions, we then apply categorization on textual contents. As a result, each category can include highly-correlated short texts aligned with the associated cognitive vector. Consequently, we extract the set of features by tracking the emotional cues in the short texts of each cognitive category. We then apply the emotion lexicons together with word vectors to learn each of the corresponding base classifiers(e.g., extraversion). To continue, given the short text contents enclosed with emotion annotations, we induce the ensemble classifier to convey emotion recognition. 
		We can aggregate the base classifiers into an ensemble to convey helpful information according to the cognitive similarities. In the \textit{online} part, we aim to predict emotion labels for the input short text. To accomplish the task, we firstly extract cognitive vectors from the input. We then select a set of relevant classifiers based on the input cognitive features. Finally, we aggregate various outputs from classifiers to make the final prediction.\\
		\vspace{-8mm}
		\section{Methodology}
		\label{methodology}
		\vspace{-1mm}
		\subsection{Offline Phase} 
		\label{offline_phase}
		\subsubsection{Inferring cognitive factors} 
		\label{Inferring-Cognitive-Factors}
		\vspace{-1mm}
		For investigating the influence of cognitive factors on emotion recognition, we need a dataset that includes both emotional and cognitive annotations. We assume that our dataset includes emotion annotations, as ground truth. Hence, we aim to compensate the cognitive annotations. To this end, we can leverage another source dataset with cognitive features to annotate our target dataset. We use a cognitive annotation dataset $D^P=\{s_i,\vec{F}_{s_i}\}(i=1, …, n)$ to infer cognitive vectors of short texts in emotional annotation dataset $D^E=\{s_i,\vec{V}_{s_i}\}(i=1, …, m)$ where $P$ and $E$ denote the source and target datasets. Here $m$ and $n$ denote the number of short texts in source and target datasets. To identify cognitive scores, we train a diverse model for each cognitive factor $q$ on $D^P$ to infer the proportionate cognitive space of $D^E$. In each model, we adopt the effective Support Vector estimation tool \cite{Awad2015} to designate a decision surface to maximize the distance between different classes. In the source dataset, there are a set of points $(s_i,f_{s_{i,j}})$ where $x_i$ is the feature vector extracted from $s_i$ and $f_{s_{i,j}}\in \vec{F}_{s_i}$ is the target value for each model $j \in [1,...,q]$. Eq. \ref{eq:objective-function-cognitive} demonstrates the objective function.\\
		\vspace{-3mm}
		\begin{equation} 
			\small
			\label{eq:objective-function-cognitive}
			\textbf{p}_q = f(x_i) = w x_i+b , b\in R, w,x_i\in R^{d}
			\vspace{-2mm}
		\end{equation}
		Where $w$ is the slope of the line, and $b$ is the intercept. Our aim is to use SVR to find a surface that minimizes the prediction error in optimization function, Eq. \ref{eq:optimization-problem-cognitive}. In regression, a soft-margin ($\epsilon$) approach is employed similar to SVM. We add slack variables $\xi_i+\xi_i^*$ to guard against outliers.\\
		\vspace{-3mm}
		\begin{equation} 
			\small
			\label{eq:optimization-problem-cognitive}
			\textbf{M}in \frac{1}{2} \Vert w\Vert^2 + C\sum_{i=1}^{n}(\xi_i+\xi_i^*)
			\vspace{-2mm}
		\end{equation}
		Here $\xi$ and $\xi^*$ are the distance of data points that lie outside the $\epsilon$ margin.
		\vspace{-3mm}
		\begin{equation} 
			\small
			\label{eq:optimization-problem-goal}
			\textbf{M}=
			\begin{cases}
				f_{s_i,j}-w^Tx_i-b\leqslant \epsilon+\xi_i\\
				w^Tx_i+b-f_{s_i,j}\leqslant \epsilon+\xi_i^*\\
				\xi_i,\xi_i^*\leqslant 0 
			\end{cases}
			\vspace{-2mm}
		\end{equation}
		Our optimization goal is to achieve the conditions in Eq. \ref{eq:optimization-problem-goal} that we solve with finding the Lagrangian in Eq. \ref{eq:Lagrangian}.\\
		\vspace{-4mm}
		\begin{equation} 
			\small
			\label{eq:Lagrangian}
			\begin{gathered}
				\mathcal{L}(w,\xi,\xi^*,\lambda,\lambda^*,\alpha,\alpha^*)=\frac{1}{2}\Vert w\Vert^2+ C\sum_{i=1}^{n}(\xi_i+\xi_i^*) +\\[-4pt]
				\sum_{i=1}^{n}\alpha^*(f_{s_i,j}-w^Tx_i-\epsilon-\xi_{i}^*)+\\[-4pt]
				\sum_{i=1}^{n}\alpha(-f_{s_i,j}+w^Tx_i-\epsilon-\xi_i)-\\[-4pt]
				\sum_{i=1}^{n}\lambda_i\xi_i+\lambda_i^*\xi_i^*
			\end{gathered}
			\vspace{-1mm}
		\end{equation}
		The Lagrange multipliers, denoted by $\lambda,\lambda^*,\alpha,\alpha^*$, are nonnegative real numbers. The minimum of Eq. \ref{eq:Lagrangian} is found by taking its partial derivatives with respect to the variables and then equating to zero. We also obtain the values of $w$ and $b$ through Eq. \ref{eq:w-cognitive} and Eq. \ref{eq:b-cognitive}. %We apply trained models in our dataset to annotating cognitive factors in our dataset.
		\vspace{-2mm}
		\begin{equation} 
			\small
			\label{eq:w-cognitive}
			\textbf{w}=\sum_{i=1}^{n}(\alpha^*-\alpha)x_i
			\vspace{-2mm}
		\end{equation}
		\begin{equation} 
			\small
			\label{eq:b-cognitive}
			\textbf{b}=-f_{s_i,j}+w^Tx_i-\epsilon
			\vspace{-3mm}
		\end{equation}
		\subsubsection{Cognitive categorization} 
		\label{Cognitive-Categorization}
		\vspace{-1mm}
		Given emotion annotated dataset $D^E$, we aim to acquire a set of cognitive categories, denoted by $C$. As elucidated in Section \ref{introductionSection}, authors with different cognitive cues express their emotions using various vocabs in emotion. Given such an intuition and based on the cognitive vector $\vec{F}_{s_i}$ of the set $P$, we can map each short text $s_i\in D^E$ to $q$ dimensions. Subsequently, we can obtain the set $C$ of cognitive categories from the emotion annotation dataset $D^E$ by splitting the space into two subspaces. Where we can define the lower and upper subspaces for each cognitive factor $p_j\in P$. Accordingly, the short texts with lower and higher levels of a cognitive factor, $p_j$, will be respectively appended to the pertinent categories of $c_{2j-1}$, $c_{2j}$. Hence, clustering \cite{Milligan1987} and border classifiers, like SVM \cite{Jain2017}, cannot cohesively distinguish the categories. To this end, we diversely adopt an entropy-based categorizing method to acquire partitioning value in each cognitive factor to obtain lower and higher bounds.
		To get the partitioning threshold vector $\vec{a}=\{\alpha_j | j\in[1,...,q]\}$, we presumed that the source dataset incorporates the cognitive factor classes. So we employed the entropy approach to attain the partition value of $\alpha_j$ for each cognitive factor $p_j$ that minimized the impurity in the resultant categories.\\
		For each given cognitive factor $p_j$, we considered a set of partitioning points in a similar range as $p_j$ in $D^P$ and evaluated them based on entropy to find the best partitioning point, $\alpha_j$. In this regard, based on each partitioning point $T$, we splitted the $D^P$ into two subsets of $d_1$ and $d_2$ and computed the entropy of resulting subsets accordingly to input cognitive class of $p_j$.
		We determined two classes $k=\{k_1,k_2\}$ for each cognitive factor $p_j\in P$ where entropy of $d_i(i=\{1,2\})$ was defined as Eq. \ref{eq:entropy-d}.\\
		\vspace{-3mm}
		\begin{equation} 
			\small
			\label{eq:entropy-d}
			Entropy(d_i)=\sum_{m\in \{1,2\}}P(k_m,d_i)\log P(k_m,d_i)
			\vspace{-2mm}
		\end{equation}
		Here $P(k_m,d_i)$ is the probability of short texts in $d_i$ pertaining class $k_m$. Given input dataset $D^P$, cognitive factor $p_j$, and partitioning point $T$, Eq. \ref{eq:entropy} computes the class information entropy $E(p_j,T,D^P)$ for the splits made by $T$. Here $\alpha_j$ is the partitioning criterion in Eq. \ref{eq:alpha-value2}.\\
		\vspace{-3mm}
		\begin{equation} 
			\small
			\label{eq:entropy}
			E(p_j,T,D^P)=\sum_{m\in \{1,2\}}\frac{|d_m|}{|D^P|}Entropy(d_m)
			\vspace{-2mm}
		\end{equation}
		\vspace{-1mm}
		\begin{equation} 
			\small
			\label{eq:alpha-value2}
			\alpha_j=\operatorname*{arg\,min}_T\{E(p_j,T,D^p)\}
			\vspace{-2mm}
		\end{equation}
		We then used $\alpha_j$ in Eq. \ref{eq:datacategory1} to obtain cognitive categories.\\
		\vspace{-3mm}
		\begin{equation} 
			\small
			\label{eq:datacategory1}
			\forall p_j\in P:
			\begin{cases}
				c_{2j-1}=\{[s_i,\vec{V}_{s_i}]|f_{i,j}< \alpha_j\}\\
				c_{2j}=\{[s_i,\vec{V}_{s_i}]|f_{i,j}\geq \alpha_j\}\\
			\end{cases}
			\vspace{-1mm}
		\end{equation}
		Consequently, the pair of sets ($c_{2j-1}$,$c_{2j}$) formed by $\alpha_j$ on each parameter $p_j$ can constitute the short texts with a low and high order of cognitive factor. We now elucidate various properties for each set of cognitive categories ($C = (c_1, ..., c_n)$):
		\vspace{-2mm}
		\begin{lemma}
			\textit{a cognitive category $c_i$ can not be empty ($\forall i\in [1, ..., 2q]: c_i\neq \varnothing$).}
		\end{lemma}
		\vspace{-3mm}
		\begin{proof}
			Since the max. and min. values for each cognitive factor $i$ do not equate ($min(D^E_i)\neq max(D^E_i)$) and the splitter parameter $\alpha_i$ is between min. and max. values ($min(D^E_i)<\alpha_i<max(D^E_i)$) so we can justify that $c_i$ can not be empty (Eq. \ref{_proof01}):
			\vspace{-2mm}
			\begin{equation}
			\label{_proof01}
				\begin{cases}
					\textit{if } \alpha_i > min(D^E_i)\Longrightarrow \exists s_j: f_{j,i}<\alpha_i\\
					\textit{if } \alpha_i < max(D^E_i)\Longrightarrow \exists s_j: f_{j,i}>\alpha_i
				\end{cases}
				\Longrightarrow c_i\neq \varnothing
			\end{equation}
		\end{proof}
		\vspace{-7mm}
		\begin{lemma}
			\textit{The aggregated data for all cognitive categories forms the original dataset $D^E$ ($\forall_i\in [1, ..., 2q]: \bigcup c_i=D^E$).}
		\end{lemma}
		\vspace{-3mm}
		\begin{proof}
			Suppose we have two categories of $c_{2j}\subseteq D^E$ and $c_{2j-1}\subseteq D^E$ where we assign the short text $s_k\in D^E$ to either $c_{2j}$ or $c_{2j-1}$. So we can justify the rules in Eq. \ref{proof_2} for the cognitive factor $j$ in $s_k$:
			\vspace{-2mm}
			\begin{equation}
				\begin{gathered}
					\forall s_k\in D^E:s_k\in c_{2j} \textit{ or } s_k\in c_{2j-1}\Longrightarrow \\[-2pt]
					c_{2j}\cup c_{2j-1}=D^E\Longrightarrow \bigcup_{i\in [1, ..., 2q]} c_i=D^E
				\end{gathered}
				\label{proof_2}
			\end{equation}
		\end{proof}
		\vspace{-8mm}
		\begin{lemma}
			\textit{The intersection of two cognitive categories for the same parameter (e.g. $c_{2j}$ and $c_{2j-1}$) partitioned by the specified threshold $\alpha_j$ will result in null. As stated in Eq. \ref{eq:Lemma3}, the intersection of the same pair ($c_{2j}$,$c_{2j-1}$) with other cognitive categories $c_i$ can be non-empty.}			
		\end{lemma}
		\vspace{-7mm}
		\begin{equation} 
			\small
			\label{eq:Lemma3}
			\begin{cases}
				\forall j\in [1, ..., q],\\
				\forall i\in [1, ..., 2q]
			\end{cases}
			\Longrightarrow
			\begin{cases}
				c_{2j-1}\cap c_i\neq \varnothing & \text{if } 2j-1\neq i\\
				c_{2j}\cap c_i \neq \varnothing & \text{if } i\neq 2j\\
				c_{2j-1}\cap c_{2j}=\varnothing\\
			\end{cases}
			\vspace{-2mm}
		\end{equation}
		\begin{proof}			
			Given cognitive factor $j$, each short text can be associated either with low $c_{2j-1}$ or high $c_{2j}$ status, resulting in $c_{2j-1}\cap c_{2j}=\varnothing$. However, given all $q$ cognitive factors for each short text with $q$ cognitive categories every pair $c_{2j-1}$ and $c_i$ can share common textual contents.
		\end{proof}
		\vspace{-5mm}
		\begin{lemma}
			 Short texts in a cognitive category $c_m$ have a similar feature according to their pertinent cognitive factor $\lfloor \frac{m+1}{2}\rfloor$. The $p_{\lfloor \frac{m+1}{2}\rfloor}$ in all of short texts are larger than $\alpha_{\lfloor \frac{m+1}{2}\rfloor}$ or are smaller than $\alpha_{\lfloor \frac{m+1}{2}\rfloor}$.\\
		\end{lemma}
		\vspace{-7mm}
		\begin{proof}
			Suppose $\forall{m\in[1,...,2q]}$ and $\forall{i,j\in [1,...,|c_m|]}$, we can use contradiction logic to prove $s_i,s_j\in c_m$ stated by $f_{i,\lfloor \frac{m+1}{2}\rfloor}$, $f_{j,\lfloor \frac{m+1}{2}\rfloor} >\alpha_m$ or $f_{i,\lfloor \frac{m+1}{2}\rfloor}$, $f_{j,\lfloor \frac{m+1}{2}\rfloor} \leq\alpha_m$. Let $s_i\in c_m$ and $s_j\notin c_m$ where $f_{i,\lfloor \frac{m+1}{2}\rfloor}$, $f_{j,\lfloor \frac{m+1}{2}\rfloor} >\alpha_m$. According to Eq. \ref{eq:datacategory1}:
			\vspace{-3mm}
			\begin{equation} 
				\begin{cases}
					\text{1. }s_i\in c_m \Rightarrow f_{i,\lfloor \frac{m+1}{2}\rfloor} > \alpha_m\\
					\text{2. }s_j \notin c_m \Rightarrow f_{j,\lfloor \frac{m+1}{2}\rfloor} <\alpha_m
				\end{cases}
			\vspace{-2mm}
			\end{equation}
			Therefore, the assumption $s_j\notin c_m$ with condition $f_{j,\lfloor \frac{m+1}{2}\rfloor} >\alpha_m$ contradicts with our initial hypothesis. Therefore, if $f_{i,\lfloor \frac{m+1}{2}\rfloor}>\alpha_m$ and $f_{j,\lfloor \frac{m+1}{2}\rfloor} >\alpha_m$, $s_i\in c_m$, $s_j$ will be in the same cognitive category($c_m$). In this way, condition $f_{i,\lfloor \frac{m+1}{2}\rfloor}$, $f_{j,\lfloor \frac{m+1}{2}\rfloor} \leq\alpha_m$ can also be proved.
		\end{proof}
		\vspace{-5mm}
		\subsubsection{Multi-channel features}
		\label{Multi-channel-features}
\begin{figure*}[t]
			\centering
			\vspace{-2mm}
			\includegraphics[width=0.65\linewidth]{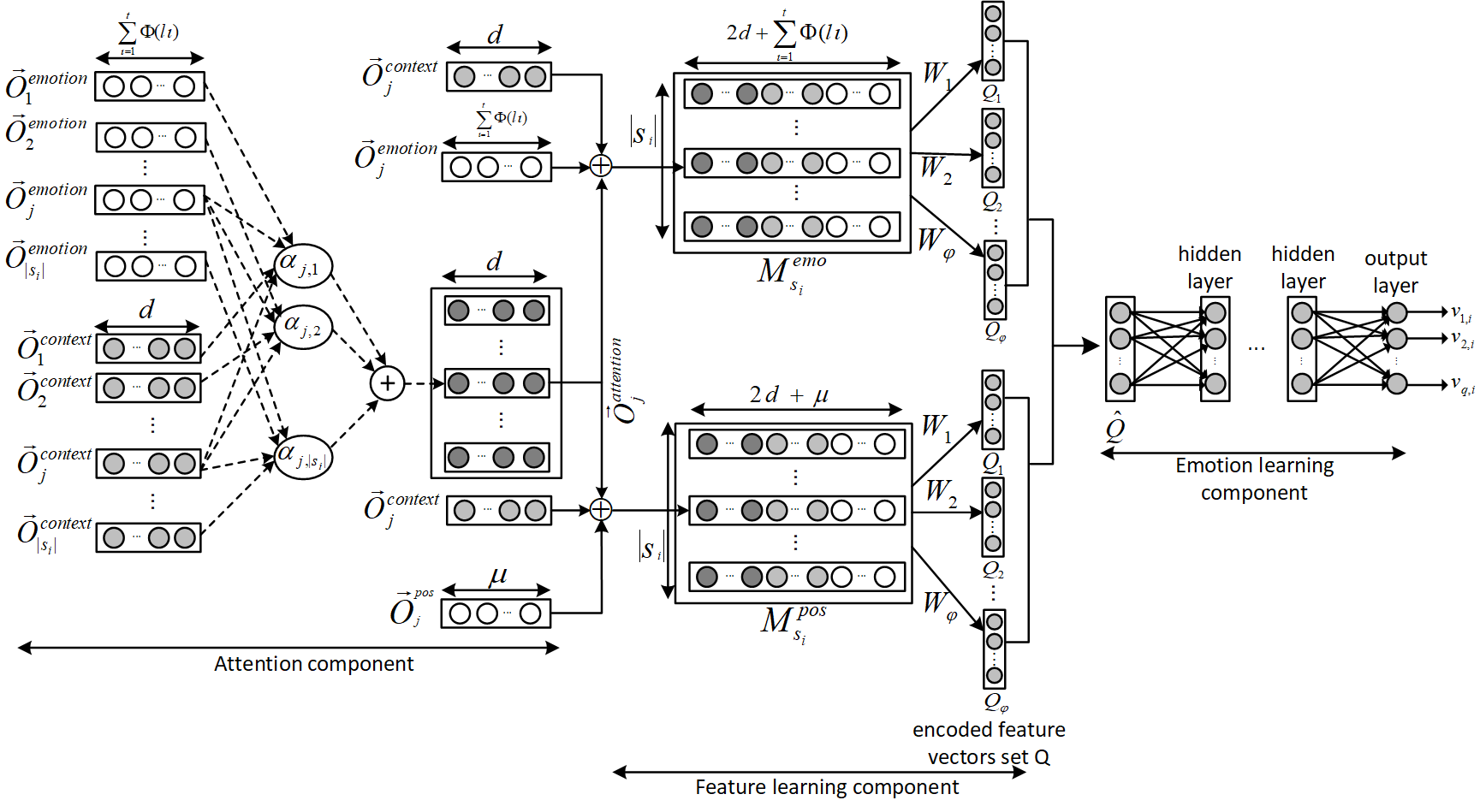}
			\vspace{-3mm}
			\caption{Base classifier architecture}
			\label{fig:base-classifier}
			\vspace{-8mm}
		\end{figure*}

		\vspace{-1mm}
		The embedding models can capture syntactic and semantic regularities within the corpus to represent each word with a real-valued vector. GloVe \cite{Hosseini2020}\cite{Pennington2014}\cite{Najafipour2020} uses word pair co-occurrences, and CBOW \cite{Mikolov2013a} predicts a word given its context. However, the resultant word vectors fail to acknowledge emotion cues in short text contents. Let $\tau$ be the corpus containing the set of words associated with textual contents of an emotion annotated dataset, $D^E$. We can then tokenize each short text $s_i$ into a set of words ($s_i=\{o_1,o_2, ...,o_{|s_i|}\}$) where $o_j \in \tau (j\in \{1, …, |s_i|\})$ is the $j^{th}$ word in short text $s_i$. We can represent the context-wise word vector corresponding to the $j^{th}$ by utilizing the embedding function $\Omega:\tau \rightarrow \mathbb{R}^d$ in Eq. \ref{eq:context-wise-word-vector}.\\ 
		\vspace{-3mm}
		\begin{equation} 
			\small
			\label{eq:context-wise-word-vector}
			\vec{O}_j^{context}=\Omega(o_j)
			\vspace{-2mm}
		\end{equation}
	 	Here, $\vec{O}_j^{context}$ is the context-wise vector for the given word $j$. Let $L=\{l_\iota|\iota\in[1,...,t]\}$ be the set and $t$ as the number of emotion lexicons where DepecheMood \cite{Araque2019} and NRC-EIL \cite{Mohammad2017a} infer the lexical vectors to designate each word with continuous scores for emotional or polarized orientations. Consequently, we can propose a hybrid vectorization process to include emotional aspects of the words. To this end, we assume that $\vec{O}_j^{emotion}= [o_{j,i}|i\in[1, ...,k]]$ is an emotion-wise word vector associated to $j^{th}$ word. Here, $k$ denotes the number of basic emotions and $o_{j,i}$ is the $i^{th}$ emotion scores for the given word $o_j$.
		\vspace{-3mm}
		\begin{equation} 
			\small
			\label{eq:emotion-wise-word-vector}
			\vec{O}_j^{emotion}=\bigoplus_{\iota=1}^{t}\Psi(o_j,l_\iota)
			\vspace{-2mm}
		\end{equation}
		As Eq. \ref{eq:emotion-wise-word-vector} formalizes, the function $\Psi: \tau\rightarrow \mathbb{R}^k$ receives the input word $o_j$ and retrieves the emotion scores through the lexicon knowledge-base $l_{\iota}$ where $t$ denotes the number of emotion lexicons and $\bigoplus$ gives the concatenation of resultant vectors. We can support $|\vec{O}_j^{emotion}|=\sum_{\iota=1}^{t}\Phi(l_\iota)$ where $\Phi$ specifies the number of emotions leveraged in $l_\iota$.\\
		Moreover, we adopt NLP processes such as POS tagging to take advantage of structural elements and syntactic patterns. Accordingly, as Eq. \ref{eq:POS-feature-vectors} elucidates, we can designate each short text with an alternative representation, POS tag-based feature vectors.\\
		\vspace{-3mm}
		\begin{equation} 
			\small
			\label{eq:POS-feature-vectors}
			\vec{O}_j^{pos}=\varrho(o_j)
			\vspace{-2mm}
		\end{equation}
		Here, $\varrho: \tau\rightarrow \mathbb{R}^\mu$ is the function that receives a word $o_j$ as input and returns the vector of the size $\mu$ with POS tags.\\
		In a nutshell, the attention-based mechanisms \cite{Zhao2016} aim to signify the words with higher impacts to foster classification procedures. As Fig. \ref{fig:base-classifier} shows, we adopt an attention module to nominate prominent focus words. Given each short text $s_i$, we can use the weighted sum of word vectors to compute the corresponding attention vector, $\vec{O}_j^{attention}$ (Eq. \ref{eq:attention-vector}).\\	
		\vspace{-3mm}
		\begin{equation} 
			\small
			\label{eq:attention-vector}
			\vec{O}_j^{attention}=\sum_{y\neq j}\alpha_{j,y}.\vec{O}_y^{context}
			\vspace{-2mm}
		\end{equation}
		As verbalized in Eq. \ref{eq:attention-weight}, $\alpha_{j,y}$ ($\alpha_{j,y}\geqslant 0$) is the attention weight subjected to $\sum_{y}\alpha_{j,y} =1$ where $"."$ denotes the element-wise multiplication.
		\vspace{-2mm}
		\begin{equation} 
			\small
			\label{eq:attention-weight}
			\begin{gathered}
				\alpha_{j,y}=\frac{1}{2}(\frac{exp(score(\vec{O}_j^{context},\vec{O}_y^{context}))}{\sum_{\acute{y}}exp(score(\vec{O}_j^{context},\vec{O}_{\acute{y}}^{context}))} +\\ \frac{exp(\varphi(\vec{O}_j^{emotion},\vec{O}_y^{emotion}))}{\sum_{\acute{y}}exp(\varphi(\vec{O}_j^{emotion},\vec{O}_{\acute{y}}^{emotion}))})
			\end{gathered}
			\vspace{-2mm}
		\end{equation}
		The $score(.,.)$ function quantifies the degree of relevance between the $j^{th}$ and $y^{th}$ words, and $\varphi$ is the similarity function that determines the correlation ratio between the word pairs according to the pertinent emotion-wise vectors. Eq. \ref{eq:score-relevance} explains how the $score(.,.)$ computes the relevance between the given pair of $o_j$ and $o_y$.\\
		\vspace{-3mm}
		\begin{equation} 
			\small
			\label{eq:score-relevance}
			score(\vec{O}_j^{context},\vec{O}_y^{context})= W_a[tanh(W_z[\vec{O}_j^{context}\oplus \vec{O}_y^{context}])]
			\vspace{-1mm}
		\end{equation}
		We randomly initialize the weights $W_a$ and $W_z$ and jointly learn them during the training process. The higher sentiment relevance between the given words in emotion classification, the larger inner-weights will be. To this end, we employ the simple but effective cosine similarity to calculate the weights between vectors, $\varphi(\vec{O}_j^{emotion},\vec{O}_y^{emotion})$.\\
		\iffalse		
		Here, the weights $W_a$ and $W_z$ are randomly initialized and jointly learned during the training process. We aim to classify short texts based on emotions which they convey. So, we assign more weight to words that are similar based on emotions which they express. To this end, we acqiure similarity between two word by utilizing cosine similarity as function $\varphi$, verbalized as Eq. \ref{eq:cosine-similarity}.
		
		\vspace{-3mm}
		\begin{equation} 
			\small
			\label{eq:cosine-similarity}		\varphi(\vec{O}_j^{emotion},\vec{O}_y^{emotion})=\frac{\vec{O}_j^{emotion}.\vec{O}_y^{emotion}}{\|\vec{O}_j^{emotion}\|\|\vec{O}_y^{emotion}\|}
			\vspace{-2mm}
		\end{equation}
	    \fi
		To attain the multichannel features, we utilize both lexicon resources and POS tags. To collectively form the multichannel features, on the one hand, we concatenate the emotion-wise and context-wise word vectors, and on the other hand, we combine the context-wise and POS-wise vector. Eq. \ref{eq:7} formulates how one can merge various vectors, including emotion-wise, context-wise, and attention vectors.
		\vspace{-2mm}
		\begin{equation} 
			\small
			\label{eq:7}
			\vec{\Upsilon}_j^{emo}= \vec{O}_j^{attention}\otimes\vec{O}_j^{context}\otimes\vec{O}_j^{emotion}
			\vspace{-2mm}
		\end{equation}
		Where the $\otimes$ is the vector concatenation operator, we can utilize Eq. \ref{eq:8} to obtain $M_{s_i}^{emo}\in \mathbb{R}^{|s_i|\times (2d+\sum_{\iota=1}^{t}\Phi(l_\iota))}$ as the matrix of vectors for $s_i$ where $|s_i|$ is the number of words.\\
		\vspace{-2mm}
		\begin{equation} 
			\small
			\label{eq:8}
			M_{s_i}^{emo} = \vec{\Upsilon}_1^{emo}\oplus \vec{\Upsilon}_2^{emo}\oplus ...\oplus\vec{\Upsilon}_{|s_i|}^{emo}
			\vspace{-2mm}
		\end{equation}
		We combine three vectors of POS, context, and attention in Eq. \ref{eq:9} to further improve the accuracy.\\
		\vspace{-3mm}
		\begin{equation} 
			\small
			\label{eq:9}
			\vec{\Upsilon}_j^{pos}= \vec{O}_j^{attention}\otimes\vec{O}_j^{context}\otimes\vec{O}_j^{pos}
			\vspace{-2mm}
		\end{equation}
		Also, Eq. \ref{eq:10} computes $M_{s_i}^{pos} \in \mathbb{R}^{|s_i|\times(2d+\mu)}$ as the vector matrix for $s_i$.\\
		\vspace{-5mm}
		\begin{equation} 
			\small
			\label{eq:10}
			M_{s_i}^{pos} = \vec{\Upsilon}_1^{pos}\oplus\vec{\Upsilon}_2^{pos}\oplus ...\oplus\vec{\Upsilon}_{|s_i|}^{pos}
			\vspace{-2mm}
		\end{equation}
		Given $s_i \in S$ comprising various number of vocabs, pertinent vectors of $M_{s_i}^{pos}$ and $M_{s_i}^{emo}$ will follow $|s_i|$. Hence, we adopt the zero padding and use $\Gamma=\operatorname*{max}_{\forall i=1,..., m}(|s_i|)$ as the fixed length for $M_{s_i}$ to unify short-text matrices.\\
		\vspace{-7mm}
		\subsubsection{Cognitive ensemble} 
		\label{Cognition-classification}
\begin{figure}[H]
			\centering
			\includegraphics[width=0.7\linewidth]{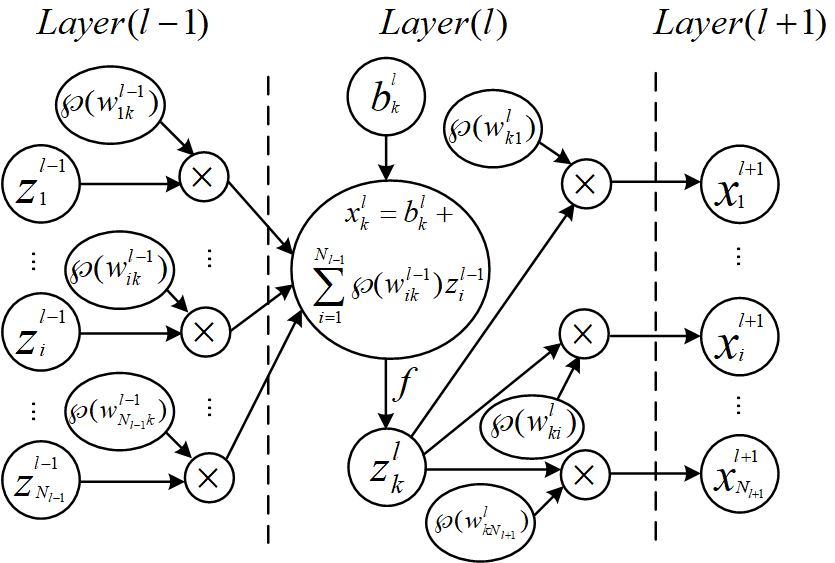}
			\vspace{-3mm}
			\caption{Input and output of a neroun at hidden layer l}
			\label{fig:Neuron}
			\vspace{-4mm}
		\end{figure}

		\vspace{-1mm}		
		Given the dataset $D^E=\{(s_i,\vec{V}_{s_i})|i=[1, ...,m]\}$ with $m$ short texts, $\vec{V}_{s_i}\in \{0,1\}^k$ can represent a binary emotion vector for $s_i$. Here, $k$ denotes the number of emotion class labels. Moreover, where the $j^{th}$ label of emotion is not null in $\vec{V}_{S_i}$, we assign $v_{i,j}$ with 1 or zero otherwise. Since $s_i$ can be concurrently involved with diverse emotions, neither single-label classification models \cite{Wei2015} nor regression algorithms \cite{Awad2015} can model such multiplexity \cite{Deng2020}. Hence, we employ multi-label classification by appointing each emotion label with a single task and adopting a parallel multi-task learner. An ensemble of convnet classifiers \cite{Wei2015} can better learn the emotion features where we designate each cognitive category $c_i$ to a distinctive multi-task classifier $h_{c_i}$. Finally, we aggregate the output of trained classifiers in $H$ (Fig. \ref{fig:base-classifier}).\\
		Given the feature learning component, we appoint the width of the filters by the dimension of word vectors, denoted by $2d+\sum_{\iota=1}^{t}\Phi(l_\iota)$. We further alter the height to acquire various sets for the encoded feature vectors. To this end, we obtain the encoded set of feature vectors $Q^{emo}=\{Q_1,Q_2,...,Q_{\varphi}\}$ for the embedding matrix $M_{s_i}^{emo}$, by $\varphi$ various window sizes, each denoted by $\delta_r\in \mathbb{N}$. Here $Q_r=[q_{r,j}|q_{r,j}\in \mathbb{R}$ and $j\in [1, ..., |s_i|+\delta_r-1]]$ represents the feature vectors for the $r^{th}$ window, Eq. \ref{eq:encode-feature-vector}.
		\vspace{-2mm}
		\begin{equation} 
			\small
			\label{eq:encode-feature-vector}
			\textbf{q}_{r,j}=\phi(M_{s_i\{j:j+\delta_r-1\}}^{emo}.W_r+b)
			\vspace{-1mm}
		\end{equation}
		$W_r\in R^{\delta_r\times (2d+\sum_{\iota=1}^{t}\Phi(l_\iota))}$ is the filter matrix, $b\in R^{\delta_r*1}$ is a bias vector, and $M_{s_i\{j:j+\delta_r-1\}}^{emo}$ is the horizontal fragment for $M_{s_i}^{emo}$ of the size $\delta_r$. The max layer consumes output feature vectors $Q^{emo}$ to exploit the final encoded vector $\hat{Q}^{emo}$ (Eq. \ref{eq:input_max_layer}).
		\vspace{-1mm}
		\begin{equation} 
			\small
			\label{eq:input_max_layer}
			\hat{Q}^{emo}=[\hat{q}_r|r\in [1, ..., \varphi]\text{ and } \hat{q}_r=\operatorname*{max}_{j\in [1,..., |s_i|-\delta_r+1]}(q_{r,j})]
			\vspace{-2mm}
		\end{equation}
		As formalized in Eq. \ref{eq:Q-hat}, we then use $\hat{Q}^{emo}$ and $\hat{Q}^{pos}$ in embedding matrix $M_{s_i}^{pos}$ to get the max layer output.\\
		\vspace{-3mm}
		\begin{equation} 
			\small
			\label{eq:Q-hat}
			\hat{Q}=\hat{Q}^{emo}\otimes\hat{Q}^{pos}
			\vspace{-2mm}
		\end{equation}
		Successively, the emotion learning component consumes $\hat{Q}$, where we collectively utilize the inter-connected layers to address the perceptual multi-label classification problem and segregate tasks in the output layer. Let $l\in\{1, .., L\}$ be the layer index of the network in the fully connected component. Given $L$ as the number of hidden layers, the index zero can determine the input for the emotion learning module. Fig. \ref{fig:Neuron} shows the schema of a neuron in the hidden layer $l$. Where $x^0$ equates to $\hat{Q}$, $x^l$ can specify the load for $l$. Similarly, $w^l$ and $z^l$ can respectively indicate the weighting matrix and the output of layer $l$, to be used in the next layer, $l+1$. Eq. \ref{eq:conv1d} attains the input of the $k^{th}$ neuron in $l$.
		\vspace{-2mm}
		\begin{equation} 
			\small
			\label{eq:conv1d}
			\textbf{x}_k^l=b_k^l+\sum_{i=1}^{N_l-1} w_{ik}^{l-1}z_i^{l-1}
			\vspace{-2mm}
		\end{equation}
		\vspace{-2mm}
		\begin{equation} 
		\small
		\label{eq:convDropout}
		\textbf{x}_k^l=b_k^l+\sum_{i=1}^{N_{l-1}} \wp(w_{ik}^{l-1})z_i^{l-1}
		\vspace{-2mm}
		\end{equation}
		Here, $x_k^l$ and $b_k^l$ can indicate the input and the bias values for the $k^{th}$ neuron in $l$. Where the output of the $i^{th}$ neuron in $l-1$ is $z_i^{l-1}$, $w_{ik}^{l-1}$ associates the weights of the $i^{th}$ neuron in $l-1$ to the $k^{th}$ neuron in $l$. Eq. \ref{eq:convDropout} neutralizes the effect of non-essential features to avoid overfitting.\\
		As explained in Sec.\ref{DropoutWeight}, the function $\wp(w_{ik}^{l-1})$ returns the weights between a pair of neurons. So as shown in Fig. \ref{fig:Neuron}, we can pass $x_k^l$ through activation function $f$, formalized in Eq. \ref{eq:relu}, to retrieve the intermediate output $z_k^l$.\\
		\vspace{-4mm}
		\begin{equation} 
			\small
			\label{eq:relu}
			\textbf{z}_k^l=max(0,x_k^l)
			\vspace{-2mm}
		\end{equation}
		\vspace{-7mm}
		\vspace{-1mm}
		\par\noindent
		The output layer constitutes $k$ output units, each dedicated to a single task. The output of the last layer in hidden layers, as the common feature representation learned for the $k$ tasks, can be fed to the output layer. The constraint can be accommodated by Eq. \ref{eq:finalOutput}, computing the network prediction output for the $j^{th}$ emotion, denoted by $\hat{v}_{j,i}\in [0,1]$.\\
		\vspace{-3mm}
		\begin{equation} 
			\small
			\label{eq:finalOutput}
			\hat{v}_{j,i}=\frac{1}{1+e^{-x_k^l}}
			\vspace{-2mm}
		\end{equation}
		To reduce the error rate, we use the back-propagation algorithm. We then employ a modified binary cross-entropy to compute the joint loss function by the predicted labels.\\
		\vspace{-4mm}
		\begin{equation} 
			\small
			\label{eq:loss-multi-label_s}
			\begin{gathered}
				\mathcal{L}=\frac{1}{m}\sum_{i\in m} (-\sum_{j=1}^{k}v_{j,i}\log(\hat{v}_{j,i})+(1-v_{j,i})\log(1-\hat{v}_{j,i}))
			\end{gathered}
			\vspace{-2mm}
		\end{equation}
		Here $\hat{v}_{j,i}$ is the output of the prediction network, $v_{j,i}$ denotes the ground-truth for $e_i\in E$, associated with the short text with index $j$. Moreover, $k$ and $m$ respectively count the numbers of emotion labels and short texts. We update the weights and bias by leveraging the loss function (Eq. \ref{eq:optimization}).
		\vspace{-2mm}
		\begin{equation} 
			\small
			\label{eq:optimization}
			w(t+1)=w(t)+\Delta w(t)
			\vspace{-2mm}
		\end{equation}
		Here, $w(t)$ and $w(t+1)$ are current weights and new weights. To compute $\Delta w(t)$, we use Adam optimizer \cite{Kingma2014} that benefits both from AdaGrad and RMSProp.\\	
		\vspace{-6mm}
				\begin{algorithm}[H]
			\vspace{0mm}
			\small
			\caption{General DropConnect}
			\label{alg_dropconnect}
			\textbf{Input:} $rate , w$\\
			\textbf{Output:} $\acute{w}$
			\begin{algorithmic}[1]
				\STATE $f=flatten(w)$ ,$\acute{w}=w$
				\STATE $l=[]$
				\WHILE{$len(l) < rate$}
				\STATE $r=randint(0,\mid f\mid)$
				\IF{not $r$ in $l$}
				\STATE $l.append(r)$
				\ENDIF
				\ENDWHILE
				\FOR {$p $ in $l$}
				\STATE $\acute{w}_{p/\mid f\mid,p\%\mid f\mid}=0$
				\ENDFOR
				\STATE return $\acute{w}$
			\end{algorithmic}
		\end{algorithm}
		\vspace{-7mm}
		\subsubsection{Weight regularization }
		\label{DropoutWeight}
		\vspace{-1mm}	

		As explained in Sec. \ref{Cognition-classification}, overfitting as a deep learning dilemma is caused by a contradiction where optimization aims to adjust the model to foster effectiveness, and generalization solely points toward inferring the unforeseen data. The dropout \cite{Srivastava2014} is the most credible regularization approach to revoke the overfitting issue. However, resolving the tension between bias and variance is not a trivial task. To this end, the dropout approaches like DropConnect \cite{Wan2013} eliminate the arbitrary weights and ignores the selected nodes in the connected layer.\\

		% Some weights increase the effect of a neuron, which in turn can result in further overfitting. To this end, the proposed procedure should eliminate the highlighted weights or alternatively reduce their impact. DropConnect randomly removes the weights that can not necessarily lead to a better-trained model, while our modified version of DropConnect learns statistics through the dataset-specific distribution and deliberately underlines the weights to better impact overfitting.
		\vspace{-6mm}
		\begin{algorithm}[H]
			\small
			\caption{None Significant Weight reduction}
			\label{alg_None_Significant_Weight}
			\textbf{Input:} $w,\alpha,\beta,\lambda$\\
			\textbf{Output:} $\acute{w}$
			\begin{algorithmic}[1]
				\STATE $f=flatten(w)$, $ \acute{w}=w$
				\STATE $\mu= \frac{1}{\mid f\mid}\sum_{i=1}^{\mid f\mid}f_i$, $ \sigma= \sqrt{\frac{1}{\mid f\mid}\sum_{i=1}^{\mid f\mid}(f_i-\mu)^2}$
				\FOR{$i$ in $\mid f\mid$}
				\STATE $r=i/\mid f\mid$ , $ c=i\%\mid f\mid$ 
				\IF{$\mu-2\sigma<f_i<\mu+2\sigma$}
				\IF{$\mu-\sigma<f_i<\mu+\sigma$}
				\STATE $\acute{w}_{r,c}=\alpha \times w_{r,c}$
				\ELSE
				\STATE $\acute{w}_{r,c}=\beta \times w_{r,c}$
				\ENDIF
				\ELSE
				\STATE $\acute{w}_{r,c}=\lambda\times w_{r,c}$
				\ENDIF
				\ENDFOR
				\STATE return $\acute{w}$
			\end{algorithmic}
		\end{algorithm}
		\vspace{-4mm}
		\par\noindent
		The DropConnect algorithm (Alg. \ref{alg_dropconnect}) turns out to be Na\"ive in the elimination process. Because it arbitrarily zeros out the selected weights. From another perspective, even the fixed dropout rate in DropConnect can reduce the model expressiveness and increase manual tuning requirements. Hence, we must initially infer the statistical cues from the embedding weights and then adjust the dropout rate consciously. To this end, we enhance the flexibility by retrieving the dropout rate based on the weights drawn from a data-specific uniform distribution.\\
		\vspace{-7mm}
		\begin{figure}[H]
			\vspace{-2mm}
			\centering
			\includegraphics[width=0.45\linewidth]{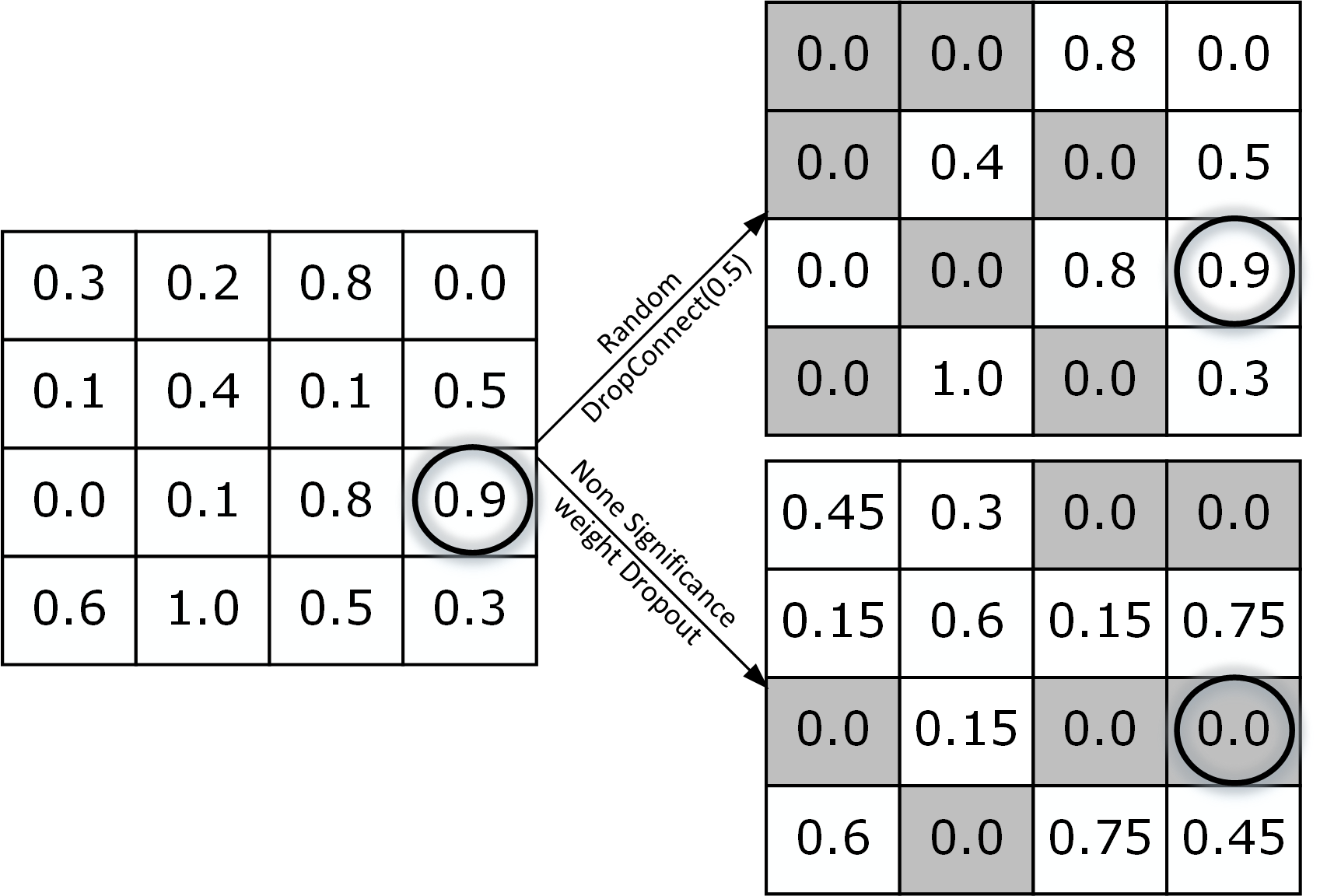}
			\vspace{-2mm}
			\caption{Random dropout versus weight regularization}
			\label{fig:Compare-Dropouts}
			\vspace{-4mm}
		\end{figure}	
		\vspace{-1mm}	
		\par
\begin{figure}[H]
		 	\centering
		 	\includegraphics[width=0.35\linewidth]{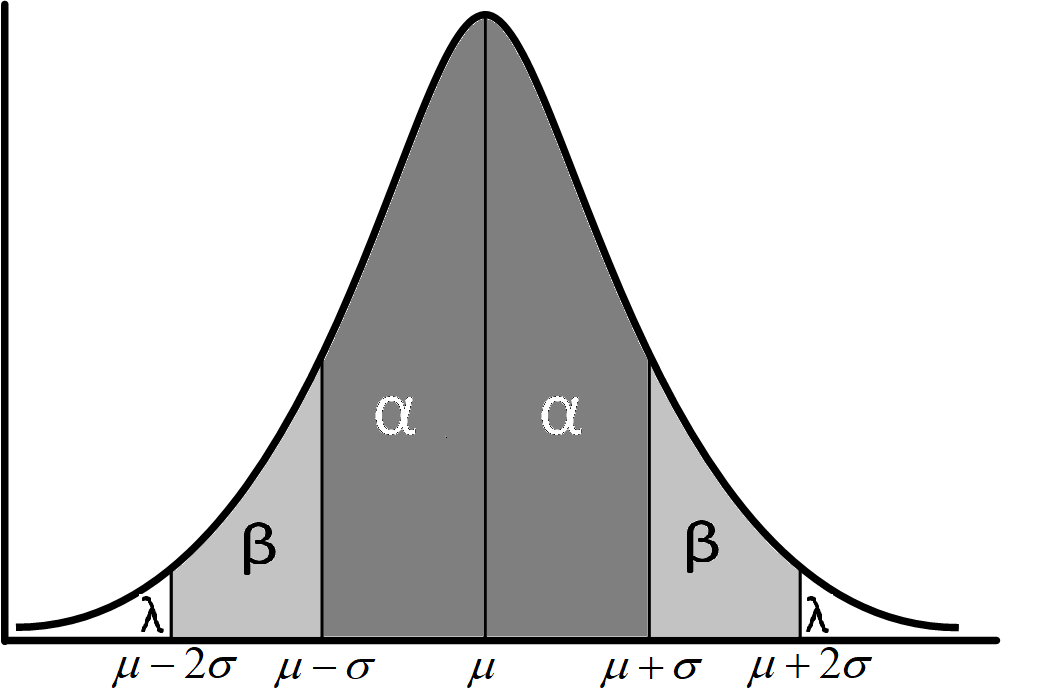}
		 	\vspace{-2mm}
		 	\caption{Distribution of coefficient weights}
		 	\label{fig:Normal-Distribution}
		 	\vspace{-4mm}
		 \end{figure}

 \noindent
		In a tractable approach, we can refer to each weight in the bell curve to initialize the elimination procedure empirically. In other words, as depicted in Fig. \ref{fig:Compare-Dropouts}, we can instantiate a weight matrix to preserve the value of the specified cell, removing or reducing the value in selected cells to adjust the activation process of the neurons. Here we can reduce the value of the given point in the matrix according to trilateral coefficients. As implemented in algorithm \ref{alg_None_Significant_Weight}, we propose an efficient dropout technique to alter imperceptive arbitrary regularization with a distribution-aware model. We multiply the outputs of neurons, highlighted in Eq. \ref{eq:convDropout}, by the justified weights based on coefficients to acquire $\acute{w}$ new weights. Given the neuron inner weights, we specify the drop-rate using dataset-oriented parameters, the standard deviation and the mean, denoted by $\mu$ and $\sigma$. As shown in Fig. \ref{fig:Normal-Distribution}, the weights are of three categories: \textit{least} ($\lambda$), \textit{minor} ($\beta$), and \textit{common} ($\alpha$). Due to the subtle connection between the significance of the neurons and the dropout, we sufficiently reduce the weights for the least and minor sections in the curve. This not only leads to a faster convergence rate but also reduces the activation weights that cause overfitting. Similarly, we relatively increase the significance of weights in the minor and common regions, coefficients of $\mu$ and $\sigma$. The changes make our model more mature through subsequent epochs, avoiding the neuron outputs to excessively rely on the least and minor weights. Finally, the model will utilize high-impact weights from the common section.
		 \vspace{-4mm}
		\vspace{-5mm}
		\subsection{Online Phase}
		\vspace{-2mm}
		In the online phase, given the cognitive vector $\vec{F}_{s_q}$ of the input short text $s_q$, the proposed method approximates the emotion vector, comprising two tasks: Detecting the cognitive factors and Estimating the emotions.\\
		\vspace{-7mm}
		\subsubsection{Cognitive factors detection}
		\label{Online-Cognitive-Factors-Detectio}
		\vspace{-1mm}
		Social media supply our propositional datasets. Hence, our framework needs to diligently handle millions of short text contents in the Online phase. To meet efficacy requirements, we train the model in Sec. \ref{Inferring-Cognitive-Factors} infrequently. Eq. \ref{eq:mdlP} can attain the cognitive vector $\vec{F}_{s_q}$ of the input query $s_q$.\\
		\vspace{-4mm}
		\begin{equation} 
			\small
			\label{eq:mdlP}
			\vec{F}_{s_q}=m(s_q) 
			\vspace{-2mm}
		\end{equation}
		\vspace{-9mm}
		\subsubsection{Cognitive aware aggregation method}
		\label{Cognitive-aware-aggregation-method}
		\vspace{-1mm}
		In this step, given the cognitive vector corresponding to the input short-text query, we firstly select a set of relevant base classifiers to perform inference as parts of the same whole. Algorithm \ref{alg:Aggregation} exemplifies how we collectively select the base classifiers and combine them accordingly.\\
		In algorithm \ref{alg:Aggregation}, we firstly extract multi-channel features out of the input query $s_q$ using the method in Sec. \ref{Multi-channel-features}. Given cognitive vector $\vec{F}_{s_q}$, we can select either of the base classifiers, $h_{2i}$ or $h_{2i-1}$. To this end, we compare each element $f_{s_q,i}\in \vec{F}_{s_q}$ to the splitter parameter $\alpha_i$ that divides the short text messages into various categories of low and high. By designating the whole corpus to the multi-task base-classifier $h_{q+1}$, we can disregard the cognitive factors in learning. We can subsequently predict the emotion vectors corresponding to $s_q$ through applying the selected base-classifiers. This results in the consensus matrix $A$ where each element $A_{i,j}$ represents the predicted class by the base classifier $j$ for an emotion $i$. Each column $j$ depicts the binary opinion of model $j$ about each of emotions in $\vec{V}_{s_q}$.\\
		\vspace{-6mm}
		\begin{algorithm}[H]
			\small
			\caption{Cognitive Aware Aggregation method}
			\label{alg:Aggregation}
			\textbf{Input:} $H,\alpha_j(j\in[1,...,q]),\vec{F}_{s_q}$\\
			\textbf{Output:} $\vec{V}_{s_q}$
			\begin{algorithmic}[1]
				\STATE $M_{s_i}^{emo}$, $M_{s_i}^{pos}=FeatureExtraction(s_q)$
				\FOR{$i$ in $[1,...,q]$}
				\IF{$f_{s_q,i}<\alpha_i$}
				\STATE $A_{:,i}=h_{2i-1}(M_{s_i}^{emo}, M_{s_i}^{pos})$
				\ELSE
				\STATE $A_{:,i}=h_{2i}(M_{s_i}^{emo}, M_{s_i}^{pos})$
				\ENDIF
				\ENDFOR
				\STATE $A_{:,q+1}=h_{q+1}(M_{s_i}^{emo}, M_{s_i}^{pos})$
				\STATE $\vec{V}_{s_q}=aggregation(A)$
				\STATE return $\vec{V}_{s_q}$
			\end{algorithmic}
		\end{algorithm}
		\vspace{-4mm}
		\par \noindent
		Consequently, we combine the base classifiers to compute the binary values from the given emotion classes, resulting in better approximation and improving the overall performance \cite{Zall2019}. To continue, we adopt the simple but effective majority voting method \cite{Zall2016} to anticipate the outcomes. Given the short text query $s_q$, Eq. \ref{eq:online-ensemble} formalizes our approach in the prediction of various sentiments.
		\iffalse
		Furthermore, we utilize the base classifiers $h_{2i-1},h_{2i}\in H$ to convey categorization using the cognitive factor $i$. Furthermore, we designate all short texts to a multi-task base classifier $h_{q+1}$ to learn generally without considering to cognitive factors. Subsequently, we can predict the emotion vectors corresponding to $s_q$ through selected base classifiers, that further form the consensus matrix $A$.\\		
		\begin{figure}[H]
			\centering
			\includegraphics[width=0.49\linewidth]{Matrix_exploiting}
			\vspace{-3mm}
			\caption{Exploiting Score}
			\label{fig:MatrixA}
			\vspace{-1mm}
		\end{figure}
		Fig. \ref{fig:MatrixA} substantiate matrix $A$ 
		\fi
		\vspace{-3mm}
		\begin{equation} 
			\small
			\label{eq:online-ensemble}
			\vec{V}_{s_q}= \bigcup_{j=1}^{k} \operatorname*{argmax}_{c_i\in \{0,1\}} (\sum_{t=0}^{q+1} g(A_{j,t},c_i))
			\vspace{-2mm}
		\end{equation}
		Eq. \ref{eq:indicator-function} shows, $A_{j,t}$ is the prediction of the base classifier $t$ using the indicator function $g(y,c)$.\\
		\vspace{-3mm}
		\begin{equation} 
			\small
			\label{eq:indicator-function}
			\textbf{g}(y,c)=\left\{ \begin{array}{rcl}
				1 & y=c\\
				0 & y\neq c\\
			\end{array}\right.
			\vspace{-2mm}
		\end{equation}
			\vspace{-8mm}
		\section{Experiment}
		\label{experiments}
		\vspace{-2mm}
		We conducted extensive experiments on multiple datasets \cite{Mohammad2018}\cite{Kosinski2013} to compare our proposed unified framework to other novel approaches in emotion detection. Taking advantage of various Python libraries and interfaces for neural networks, we ran the experiments on a server with a 4.20 GHz Intel Core i7-7700K CPU and 64GB of RAM. The codes are available to download \footnote{\url{https://sites.google.com/view/EmoDNN}}.
		\vspace{-6mm}
		\subsection{Data}
		\label{dataset}
		\vspace{-2mm}
		We used three datasets to examine our method in detecting personality and emotions from brief contents.\\
        - $MyPersonality$ \cite{Kosinski2013}: The MyPersonality dataset, denoted by $D^P$, predicts the cognitive labels for our target emotion dataset \cite{Mohammad2018} and comprises the cues for extraversion, agreeableness, conscientiousness, and neuroticism. We eliminate the effect of openness due to minor significance.\\
        - $SemEval2018$ \cite{Mohammad2018}: This dataset ($D^E$) is annotated by 11 emotion tags. Like Ekman’s standard \cite{Ekman1999}, we include fear, anger, joy, disgust, and sadness.\\
        - $WASSA-2017$ \cite{Mohammad2017}: is the destination dataset, denoted by $D^E$, and includes fear, joy, sadness, and anger emotions. We utilize this emotion annotated dataset to evaluate the performance of our proposed framework in multi-class labeling.
		We enclose the statistics pertaining MyPersonality and SemEval2018 in Tables \ref{tab:Statistics-MyPersonality} and \ref{tab:Statistics-SemEval2018}, respectively.\\
		\vspace{-6mm}
		\begin{table}[H]
			\centering
			\def\arraystretch{1.5}
			\tiny
			\begin{tabular}{|c|c|c|c|c|c|c|c|l|}
				\hline
				& \multicolumn{2}{c|}{NEU}  & \multicolumn{2}{c|}{CON}  & \multicolumn{2}{c|}{EXT}  & \multicolumn{2}{c|}{AGR}  \\ \hline
				& Low          & High       & Low          & High       & Low          & High       & Low          & High       \\ \hline
				\#Tweet & 6200         & 3717       & 5361         & 4556       & 5707         & 4210       & 4649         & 5268       \\ \hline
				Max    & \multicolumn{2}{c|}{4.75}    & \multicolumn{2}{c|}{5}    & \multicolumn{2}{c|}{5}    & \multicolumn{2}{c|}{5}    \\ \hline
				Min    & \multicolumn{2}{c|}{1.25} & \multicolumn{2}{c|}{1.45} & \multicolumn{2}{c|}{1.33} & \multicolumn{2}{c|}{1.65} \\ \hline
				Mean    & \multicolumn{2}{c|}{2.6} & \multicolumn{2}{c|}{3.47} & \multicolumn{2}{c|}{3.35} & \multicolumn{2}{c|}{3.62} \\ \hline
				STD    & \multicolumn{2}{c|}{0.76} & \multicolumn{2}{c|}{0.74} & \multicolumn{2}{c|}{0.85} & \multicolumn{2}{c|}{0.68} \\ \hline
				Median    & \multicolumn{2}{c|}{2.6} & \multicolumn{2}{c|}{3.4} & \multicolumn{2}{c|}{3.4} & \multicolumn{2}{c|}{3.65} \\ \hline
			\end{tabular}
			\vspace{-2mm}
			\caption{Statistics pertaining MyPersonality dataset}
			\label{tab:Statistics-MyPersonality}
			\vspace{-6mm}
		\end{table}
		\begin{table}[H]
			\centering
			\def\arraystretch{1.5}
			\tiny
			\begin{tabular}{|c|c|c|c|c|c|c|c|c|c|c|c|c|c|}
				\hline
				& Anger & Disgust & Fear & Joy & Sadness \\ \hline
				\#Tweet & 2859   & 2921   & 1363  & 2877     & 2273   \\ \hline
				Max    & 1       & 1      & 1        & 1          & 1        \\ \hline
				Min    & 0    & 0     & 0   & 0     & 0      \\ \hline
				AVG    & 0.37                    & 0.378                      & 0.18                   & 0.37                  & 0.2943                      \\ \hline
				STD    & 0.48                   & 0.485                    & 0.38                  & 0.48                 & 0.455                    \\ \hline
				Median    & 0.0                   & 0.0                    & 0.0                  & 0.0                 & 0.0                    \\ \hline
			\end{tabular}
			\vspace{-2mm}
			\caption{Statistics of SemEval2018 dataset}
			\label{tab:Statistics-SemEval2018}
			\vspace{-5mm}
		\end{table}
		\par
\begin{figure}[H]
			\begin{subfigure}{.45\linewidth}
				\centering
				% include first image
				\includegraphics[width=1\linewidth]{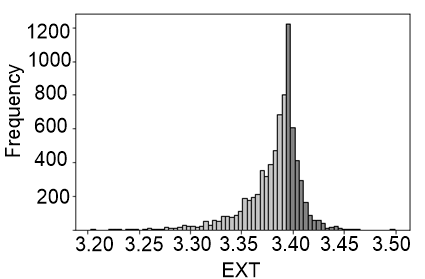}
				\caption{EXT}
				\label{fig:bar-EXT1}
			\end{subfigure}
			\begin{subfigure}{.45\linewidth}
				\centering
				% include second image
				\includegraphics[width=1\linewidth]{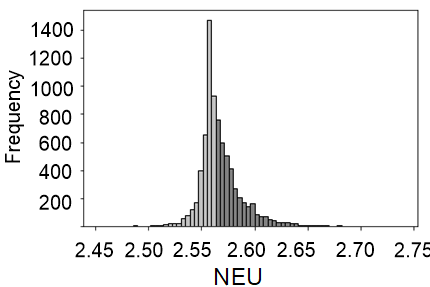}
				\caption{NEU}
				\label{fig:bar-NEU1}
			\end{subfigure}
			\newline
			\begin{subfigure}{.45\linewidth}
				\centering
				% include third image
				\includegraphics[width=1\linewidth]{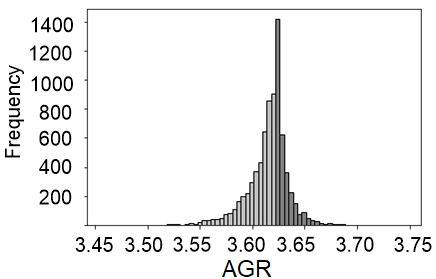}
				\caption{AGR}
				\label{fig:bar-AGR1}
			\end{subfigure}
			\begin{subfigure}{.45\linewidth}
				\centering
				% include fourth image
				\includegraphics[width=1\linewidth]{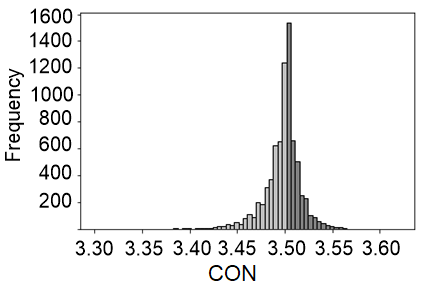}
				\caption{CON}
				\label{fig:bar-CON1}
			\end{subfigure}
			\vspace{-3mm}
			\caption{Cognitive factors data distribution}
			\label{fig:distribution-cognitive}
		\end{figure}

 \noindent Fig. \ref{fig:distribution-cognitive} shows short text distribution for each given cognitive factor, where the $x$ axis reports the weights and $y$ counts the frequency. The middle threshold differentiates the low and high domains with respective light and dark colors.\\
		\vspace{-8mm}
		\vspace{-4mm}
		\iffalse
		\vspace{4mm}
		Fig. \ref{fig:data-emotion} depicts the distribution of the tweets given the emotions expressed within contents. As observed, compared to Joy, Disgust, and anger, microblog authors are reluctant to reveal other private emotions such as fear and sadness.\\
		\iffalse
		Fig. \ref{fig:data-emotion} designate the distributions of tweets based on emotions which are expresses. As you can see, the emotion of fear has the least number of tweets. The distributions of tweets according to each cognitive factor is depicted in Fig. \ref{fig:distribution-cognitive}. As can be seen, the data have approximately the same distribution in each cognitive category excepted d extroversion.\\
		\fi
		\vspace{-6mm}
		\begin{figure}[H]
			\centering
			\includegraphics[width=0.6\linewidth]{data_emotion_Distribution}
			\vspace{-3mm}
			\caption{Emotion Data Distribution}
			\label{fig:data-emotion}
			\vspace{-1mm}
		\end{figure}
		\fi			
		\iffalse
		Our Twitter dataset \cite{Hua2017} includes 8 million English tweets in Australia, collected via \textit{Spritzer Twitter Feed}. The sampling was done at various times of the day for a complete year. We then used \textit{Twitter API} to select approximately $4K$ users from streaming tweets and retrieved up to 1000 records from their Twitter history. Finally, we attained $\approx 1M$ geo-tagged \textit{tweets} which are all composed in Australian territory. The dataset contains 305K vocabs and is made of 65M collocations.
		\fi
		\vspace{-6mm}
		\subsection{Benchmark}
		\label{Benchmark}
		\vspace{-1mm}
		Intuitively, we define hypothetic parameters to evaluate the effectiveness of our proposed framework in emotion recognition. The statistical parameters are as follows: True positive is observed when a short text has emotion $e_i$ and the model predicts the same. False positive indicates that the short-text doesn't relate to the emotion $e_i$ but the model predicts oppositely. Applying similar logic can determine True and False Negatives. We can calculate the metrics of Accuracy, Precision, Recall, and F-measure using TP, FP, TN, and FN. Accordingly, we can distinguish the best performance by F-measure while we apply 10-fold cross-validation in every evaluation process.\\		
		\iffalse
		Accuracy in Eq.\ref{eq:Accuracy} is a metric that quantifies the number of correct predictions made out of all predictions.
		\begin{equation} 
			\small
			\label{eq:Accuracy}
			Accuracy=\frac{TP+TN}{TP+TN+FN+FP}
			\vspace{-2mm}
		\end{equation}
		Recall in Eq.\ref{eq:Recall} is a metric that quantifies the number of correct positive predictions made out of all positive predictions that could have been made.
		\begin{equation} 
			\small
			\label{eq:Recall}
			Recall=\frac{TP}{TP+FN}
			\vspace{-2mm}
		\end{equation}
		Precision in Eq.\ref{eq:Precision} is a metric that quantifies the number of correct positive predictions made.
		\begin{equation} 
			\small
			\label{eq:Precision}
			Precision=\frac{TP}{TP+FP}
			\vspace{-2mm}
		\end{equation}
		F-Measure in Eq.\ref{eq:F-Measure} provides a way to combine both precision and recall into a single measure that captures both properties.
		\begin{equation} 
			\small
			\label{eq:F-Measure}
			F1=2*\frac{Precision*Recall}{Precision+Recall}=\frac{2TP}{2TP+FP+FN}
			\vspace{-2mm}
		\end{equation}
		\fi
		\vspace{-9mm}
		\subsection{Baselines}
		\label{baselines}
		\vspace{-1mm}
		We employ the benchmark in Sec. \ref{Benchmark} to examine the performance of the rival methods in emotion recognition:
		\vspace{-2mm}
		\begin{itemize}
			\item $Uunison$:
			This baseline \cite{Colneric2018} leverages a variety of deep learning modules, such as word and character-based RNN and CNN, to improve traditional classifiers including BOW and latent semantic indexing.
			\item $Senti_{HC}$:
			This model is a hierarchical classification scheme that comprises three levels in the learning process: neutrality(neutrality versus emotionality), polarity, and emotions(five basic emotions) \cite{Esmin2012}.
			\item $SVM-Behavior$:
			Similar to \cite{Jain2017}, it combines unigrams and emotion lexicons and uses SVM-Behavior to classify text contents according to emotion cues.
			\item $lexicon based$:
			Instead of word embedding\cite{Araque2019}, this model is performed by emotion lexicon.
			\item $EmoDNN_{SVM}$:
			This model is based on our proposed categorization method, but the learning component employs an SVM classifier on unigrams.
			\item $EmoDNN_{wd}$: This method replaces multichannel feature learning with text embedding \cite{Hosseini2020}\cite{Pennington2014}.
			\item $EmoDNN$:
			Our proposed framework in Sec. \ref{Framework_Overview}.
			\vspace{-4mm}		
		\end{itemize}
		\subsection{Effectiveness}
		\vspace{-1mm}
		\subsubsection{Impact of learning parameters on emotion recognition}
		\vspace{-1mm}		
		\label{Effect-Number-Epochs-Mini-Batches}
		Given the importance of the batch size is in the dynamics of deep learning algorithms, we designate this section to measure the accuracy for each given emotion where the batch size varies. Table \ref{tab:Accuracy-in-batch-size} shows where the batch size varies the accuracy fluctuates up to 5\% with minimum and maximum for Disgust and Joy emotions. As a result, we select the best value of 128 tweets for the batch size in our method to maximize the performance. Similarly, we have attained the best batch size for other rivals. Similarly, Table \ref{tab:Accuracy-in-epoch-size} investigates the impact of the number of epochs on the accuracy. Excluding the fear and sadness emotions, where the epoch is set to 50, we gain the best effectiveness.\\
		Furthermore, we need to evaluate the embedding module. Hence, Table \ref{tab:Accuracy-in-glove-size} reports the accuracy for various embedding dimensions where we opt for the value of 200 to get the best overall performance. In retrospect, the lower dimensions can better adjust to fewer data, like for fear and sadness. Because the higher the dimension in low sampling, the bigger the data sparsity will be.\\
		\vspace{-6mm}
		\begin{table}[H]
			\centering
			\def\arraystretch{1.5}
			\tiny
			\begin{tabular}{|c|c|c|c|c|c|}
				\hline
				\multirow{2}{*}{\begin{tabular}[c]{@{}c@{}}batch \\ size\end{tabular}} & \multicolumn{5}{c|}{Accuracy}             \\ \cline{2-6} 
				& Anger & Disgust & Fear  & Joy   & Sadness \\ \hline
				30                                                                     & 79.85 & 77.44   & \textbf{92.32} & 77.28 & \textbf{82.33}   \\ \hline
				50                                                                     & 80.06 & 76.45   & 92.24 & 78.01 & 81.85   \\ \hline
				80                                                                     & 80.4  & 76.9    & 91.99 & 77.42 & 81.54   \\ \hline
				100                                                                    & 80.16 & 77.29   & 91.77 & 77.91 & 81.98   \\ \hline
				128                                                                    & \textbf{81.73} & \textbf{77.67}   & 89.39 & \textbf{83.05} & 78.1    \\ \hline
			\end{tabular}
			\vspace{-2mm}
			\caption{Impact of batch-size on accuracy}
			\label{tab:Accuracy-in-batch-size}
			\vspace{-3mm}
		\end{table}
		\vspace{-3mm}
		\begin{table}[H]
			\centering
			\def\arraystretch{1.5}
			\tiny
			\begin{tabular}{|c|c|c|c|c|c|}
				\hline
				\multirow{2}{*}{epoch} & \multicolumn{5}{c|}{Accuracy}             \\ \cline{2-6} 
				& Anger & Disgust & Fear  & Joy   & Sadness \\ \hline
				40                     & 80.51 & 76.67   & 92.01 & 77.27 & 81.7    \\ \hline
				50                    & \textbf{81.73} & \textbf{77.67}   & 89.39 & \textbf{83.05} & 78.1    \\ \hline
				80                     & 80.32 & 76.27   & \textbf{92.02} & 77.49 & 81.53   \\ \hline
				100                     & 80.35 & 76.79   & 91.76 & 77.79 & \textbf{81.88}   \\ \hline
			\end{tabular}
			\vspace{-2mm}
			\caption{Impact of number of epoch on accuracy}
			\label{tab:Accuracy-in-epoch-size}
			\vspace{-3mm}
		\end{table}
		\vspace{-3mm}
		\begin{table}[H]
			\centering
			\def\arraystretch{1.5}
			\tiny
			\begin{tabular}{|l|l|l|l|l|l|}
				\hline
				\multicolumn{1}{|c|}{\multirow{2}{*}{glove}} & \multicolumn{5}{c|}{Accuracy}                                                                                                                   \\ \cline{2-6} 
				\multicolumn{1}{|c|}{}                       & \multicolumn{1}{c|}{Anger} & \multicolumn{1}{c|}{Disgust} & \multicolumn{1}{c|}{Fear} & \multicolumn{1}{c|}{Joy} & \multicolumn{1}{c|}{Sadness} \\ \hline
				25                                           & 79.55                      & 77.07                        & 91.12                     & 74.64                    & 80.58                        \\ \hline
				50                                           & 81.49                      & 76.29                        & 91.77                     & 73.29                    & \textbf{82.22}                        \\ \hline
				100                                          & 79.96                      & 74.92                        & \textbf{92}                        & 76.97                    & 82.18                        \\ \hline
				200                                          & \textbf{81.73}                      & \textbf{77.67}                        & 89.39                     & \textbf{83.05}                    & 78.1                         \\ \hline
			\end{tabular}
			\vspace{-2mm}
			\caption{Impact of text-embedding on accuracy}
			\label{tab:Accuracy-in-glove-size}
			\vspace{-4mm}
		\end{table}
		\vspace{-1mm}
		\par \noindent
		The learning rate parameter can significantly affect the robustness of the proposed model as it can directly influence the optimization weights. As for larger learning rates, the chance to exceed the extreme point will be bigger, causing an unstable system. Conversely, where the learning rate decreases, the training time can exquisitely take longer. As observed in Table \ref{tab:Accuracy-in-learning-rate}, the learning rate of $10^{-6}$ results in the highest accuracies in the majority of emotions.\\		
		\vspace{-6mm}
		\begin{table}[H]
			\centering
			\def\arraystretch{1.5}
			\tiny
			\begin{tabular}{|c|c|c|c|c|c|}
				\hline
				\multirow{2}{*}{\begin{tabular}[c]{@{}c@{}}learning \\ rate\end{tabular}} & \multicolumn{5}{c|}{Accuracy}                \\ \cline{2-6} 
				& Anger  & Disgust & Fear   & Joy    & Sadness \\ \hline
				$10^{-1}$                                                                       & 79.86  & 75.74   & 91.51  & 76.65  & 80.28   \\ \hline
				$10^{-2}$                                                                      & 79.81  & 77.1    & \textbf{92.03}  & 76.16  & 81.82   \\ \hline
				$10^{-3}$                                                                     & 79.48  & 75.94   & 91.91  & 79.33  & 80.86   \\ \hline
				$10^{-4}$                                                                    & 80.41  & \textbf{76.94}   & 91.2   & 77.68  & \textbf{81.84}   \\ \hline
				$10^{-5}$                                                                   & 81.11 & 77.15  & 89.49 & 82.97 & 76.7   \\ \hline
				$10^{-6}$                                                                  & \textbf{81.73}  & \textbf{77.67}   & 89.39  & \textbf{83.05}  & 78.1    \\ \hline
			\end{tabular}		
			\vspace{-2mm}
			\caption{Impact of learning rateg on accuracy}
			\label{tab:Accuracy-in-learning-rate}
			\vspace{-4mm}
		\end{table}
		\begin{table*}[]
			\vspace{-2mm}
			\centering
			\def\arraystretch{1.5}
			\tiny
			\begin{tabular}{|c|c|c|c|c|c|c|c|c|c|c|c|c|c|c|c|}
				\hline
				\multirow{2}{*}{methods} & \multicolumn{3}{c|}{Anger}                   & \multicolumn{3}{c|}{Disgust}                & \multicolumn{3}{c|}{Fear}                     & \multicolumn{3}{c|}{Joy}                      & \multicolumn{3}{c|}{Sadness}                  \\ \cline{2-16} 
				& Percision     & Recall       & F1            & Percision     & Recall       & F1           & Percision     & Recall        & F1            & Percision     & Recall        & F1            & Percision     & Recall        & F1            \\ \hline
				Unison                   & \textbf{0/84} & 0/3          & 0/41          & \textbf{0/93} & 0/27         & 0/42         & \textbf{0/84} & 0/29          & 0/42          & 0/8           & 0/68          & 0/73          & \textbf{0/84} & 0/23          & 0/34          \\ \hline
				Senti\_\{HC\}            & 0/54          & 0/54         & 0/53          & 0/53          & 0/52         & 0/52         & 0/66          & 0/66          & 0/66          & 0/5           & 0/58          & 0/5           & 0/55          & 0/56          & 0/55          \\ \hline
				SVM\_Behavior            & 0/68          & 0/68         & 0/68          & 0/65          & 0/66         & 0/65         & 0/72           & \textbf{0/78} & \textbf{0/75} & 0/68          & 0/69          & 0/68          & 0/52          & \textbf{0/71} & 0/6 \\ \hline
				Lexicon-based            & 0/33          & 0/4          & 0/36          & 0/34          & 0/4          & 0/37         & 0/54          & 0/6           & 0/56          & 0/33          & 0/48          & 0/39          & 0/6           & 0/4           & 0/48          \\ \hline
				EmoDNN\_\{SVM\}         & 0/77          & 0/61         & 0/68          & 0/7           & 0/57         & 0/63         & 0/78          & 0/4           & 0/53          & 0/81          & 0/63          & 0/71          & 0/68          & 0/41          & 0/51          \\ \hline
				EmoDNN\_\{wd\}          & 0/7           & \textbf{0/8} & 0/74          & 0/77          & 0/63         & 0/69         & 0/68          & 0/64          & 0/65          & 0/75          & \textbf{0/75} & \textbf{0/75} & 0/62          & 0/53          & 0/57          \\ \hline
				EmoDNN                  & 0/75          & 0/75         & \textbf{0/75} & 0/7           & \textbf{0/7} & \textbf{0/7} & 0/75          & 0/6           & 0/67          & \textbf{0/83} & 0/69          & \textbf{0/75} & 0/64          & 0/61          & \textbf{0/62}          \\ \hline
			\end{tabular}
			\vspace{-2mm}
			\caption{Precision, Recall, and F-measure(F1) of Each Emotion in Different Methods}
			\label{tab:F1-in-different-methods}
			\vspace{-8mm}
		\end{table*}
		\vspace{-3mm}
		\subsubsection{Effectiveness of EmoDNN in multi-label dataset}
\begin{figure}[H]
			\vspace{-7mm}
			\centering
			\includegraphics[width=0.60\linewidth]{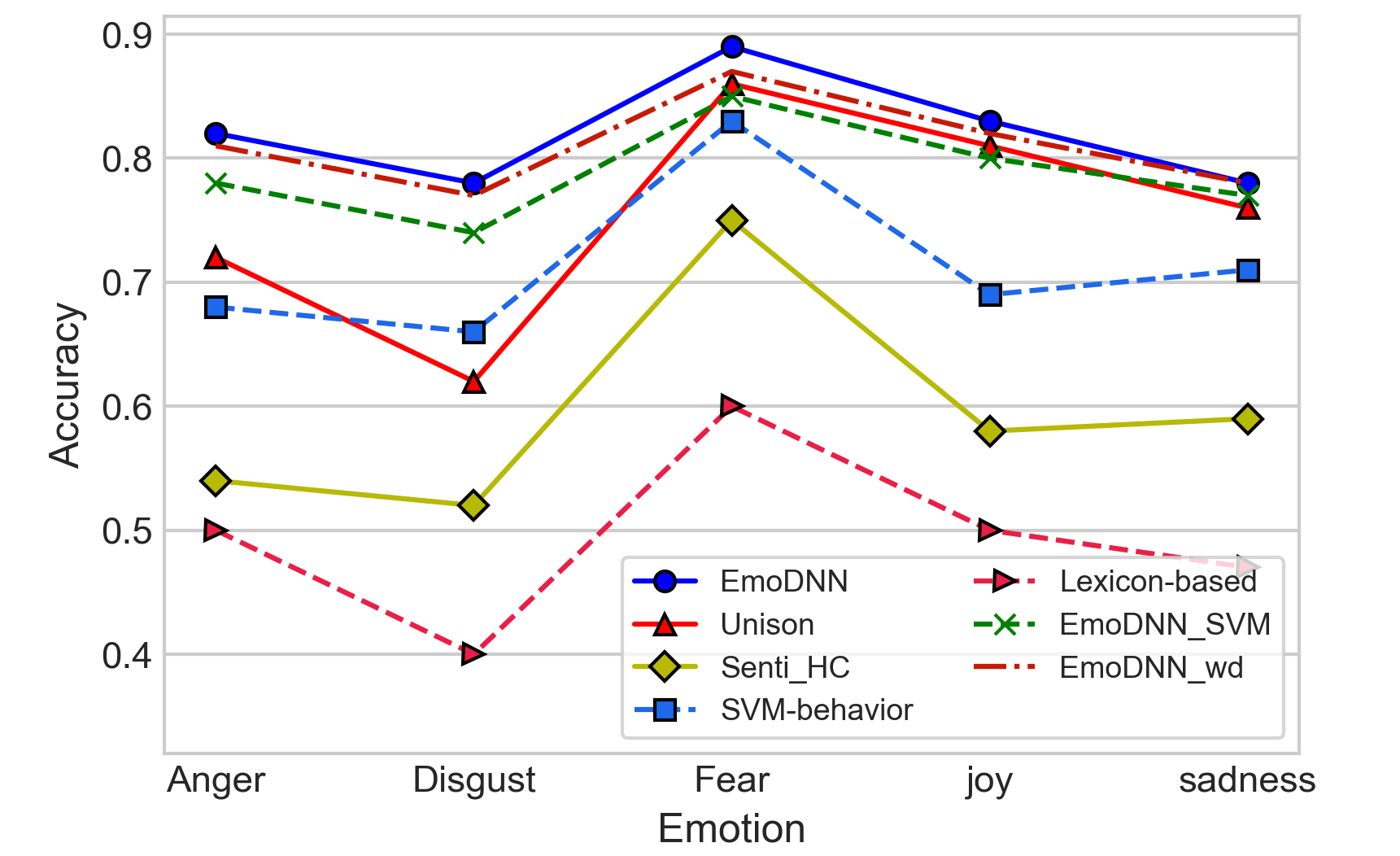}
			\vspace{-3mm}
			\caption{Compare baselines accuracy}
			\label{fig:compare-baselines-accuracy}
			\vspace{-4mm}
		\end{figure}

		\vspace{-1mm}		
		\label{Comparison-with-Traditional-Emotion-Detection}	
		We employ the benchmark (Sec. \ref{Benchmark}) to compare the rivals (Section \ref{baselines}) in inferring the emotions from brief contents. We observe in Fig. \ref{fig:compare-baselines-accuracy} that the performance of all the methods is more than 40\% which is even better for the neural network models. However, both versions of our proposed approach, including $EmoDNN_{SVM}$ and $EmoDNN_{wd}$, turn out to be the best classifiers with an improvement of up to 6.6\% versus the best performing competitor, Unison. From another perspective, lack of training procedure justifies why the lexicon-based methods attain the lowest accuracy. Even though we integrated our framework with shallow machine learning methods, e.g., SVM, the modified solution was still capable of overcome other baselines, where applying deep learning modules assured better accuracy. To prevent overfitting, we introduced a new improved dropout mechanism to foster the classification task, with further improvement of 1\% compared to the arbitrary dropout. We also adopted the values 1.5, 1, and 0, for $\alpha$,$\beta$, and $\lambda$ coefficients and utilized a modified emotion-aware embedding approach instead of a pre-trained vector module, improving the accuracy by up to 1.07\%.\\
		\par \noindent
		Table \ref{tab:F1-in-different-methods} compares the performance based on emotions via 10-fold cross-validation where our cognitive-aware emotion detection approach overpasses other baselines. The reason is three-fold: We include cognitive inference, enhance the dropout, and equip the feature vectors with emotion-aware cues, making EmoDNN gain high F1-measure values of 0.75, 0.70, and 0.75 for anger, disgust, and joy.\\
		\begin{figure}[H]
			\centering
			\vspace{-7mm}
			\includegraphics[width=0.60\linewidth]{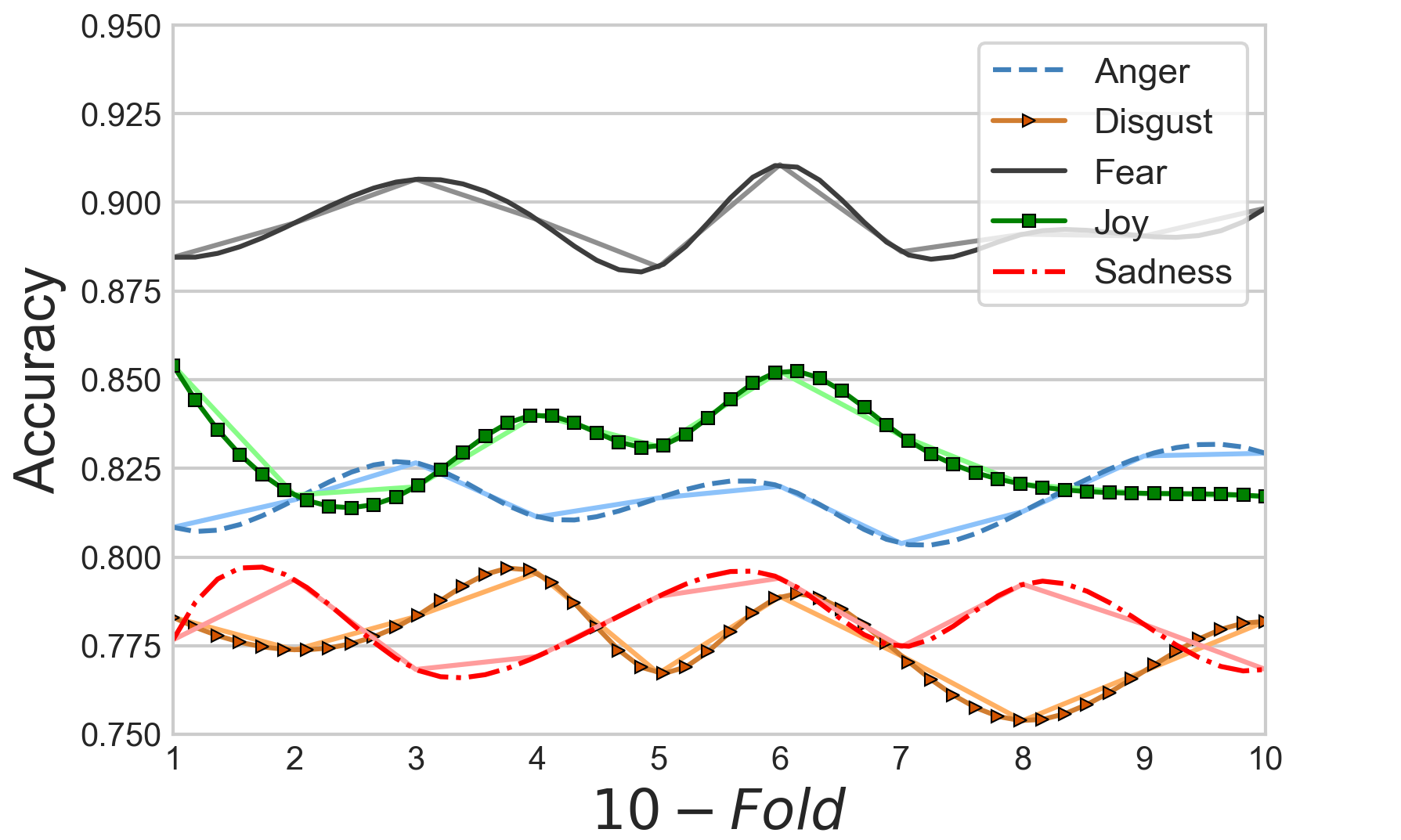}
			\vspace{-3mm}
			\caption{Compare accuracy between emotions}
			\label{fig:compare-emotions}
			\vspace{-5mm}
		\end{figure}

		\par
\begin{figure}[H]
			\centering
			\vspace{-4mm}
			\begin{subfigure}{0.45\linewidth}
				\centering
				\includegraphics[width=1\linewidth]{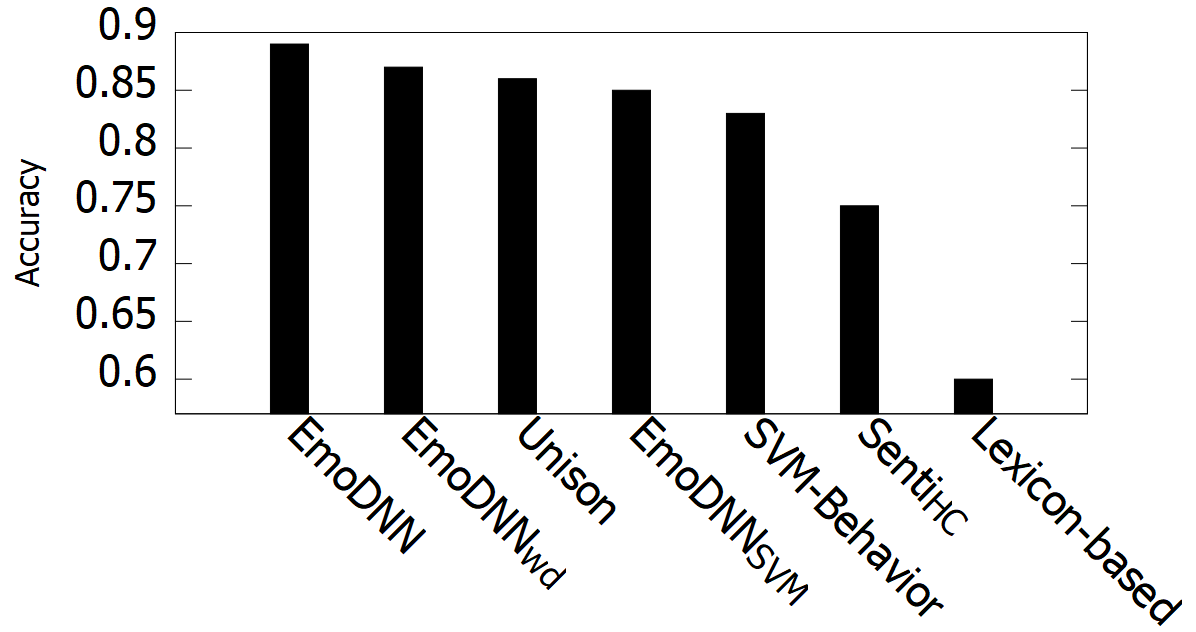}
				\vspace{-7mm}
				\caption{Fear}
				\label{fig:bar-acc-fear}
			\end{subfigure}
			\centering
			\begin{subfigure}{0.45\linewidth}
				\centering
				\includegraphics[width=1\linewidth]{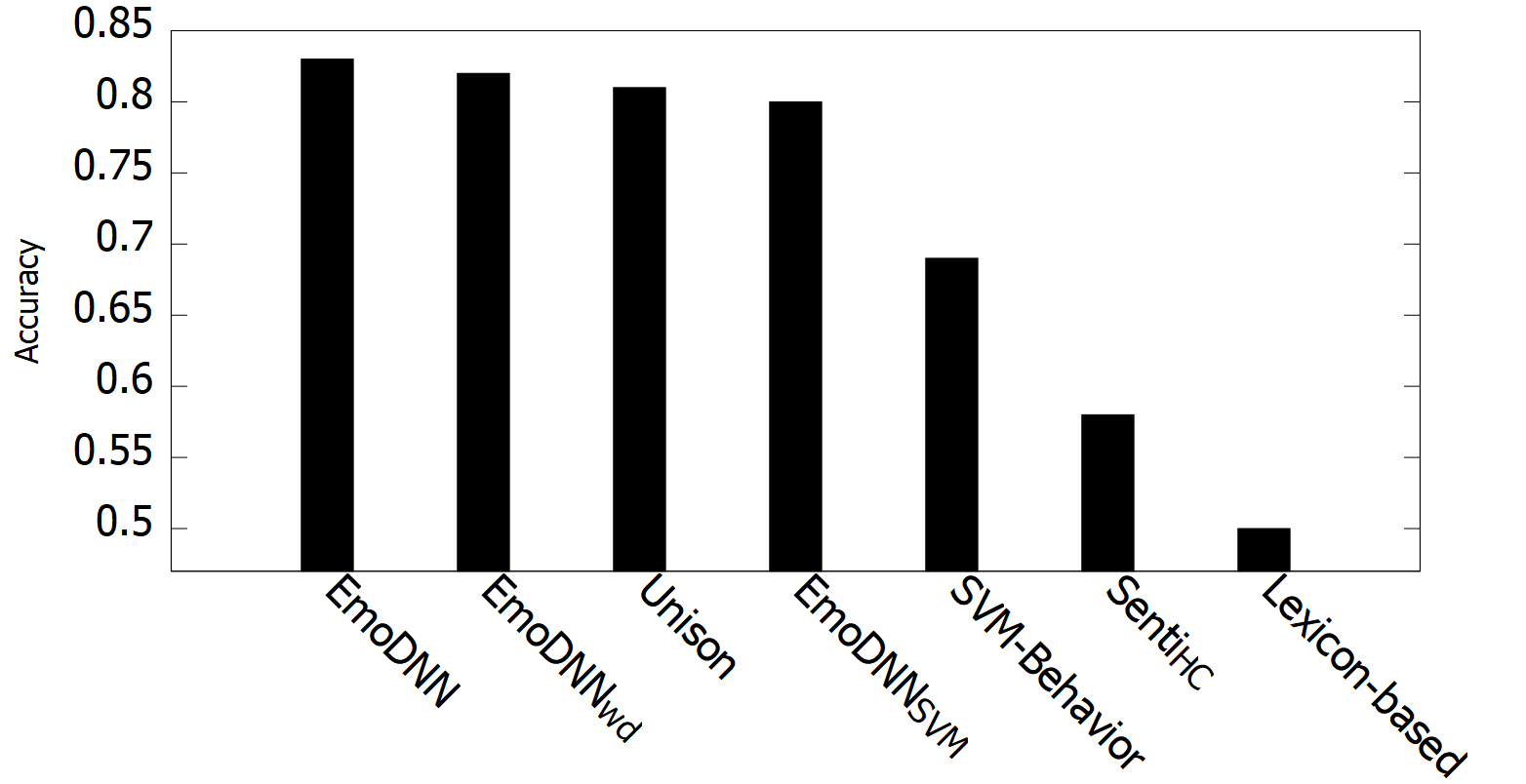}
				\vspace{-7mm}
				\caption{Joy}
				\label{fig:bar-acc-joy}
			\end{subfigure}
			\hfill
			\vspace{-3mm}
			\caption{Compare baselines accuracy in fear and joy emotion}
			\label{fig:bar-acc}
			\vspace{-3mm}
		\end{figure}

 \noindent
		Aiming to test out-of-samples in various folds, we study how the accuracy of our proposed method fluctuates in different emotions. We observe (Fig. \ref{fig:compare-emotions}) that fear and disgust gain the best and the least accuracies. While recognition of fear and joy is convenient, the detection of sadness and disgust is tedious in brief contents. We leverage the intuition in Fig. \ref{fig:bar-acc} to compare the effect of the two highest accuracies based on fear and joy where EmoDNN surpasses other rivals and the lexicon-based attains the least accuracy.
		\vspace{-4mm}
		\subsubsection{Effectiveness of EmoDNN in multi-class dataset}
		\vspace{-1mm}
		We choose the multi-class WASSA-2017 \cite{Mohammad2017} emotion recognition dataset to evaluate mutual co-existed labels. We compare our method with unison and SVM-behavior methods that address the multi-class challenge. Table \ref{tab:MultiClass_compare} shows that EmoDNN overcomes both competitors and unison, equipped with DL modules, can overpass SVM classifiers. Since we include the cognitive cues in emotion features, EmoDNN can upgrade unison by up to 2\% in F1-measure.
		\begin{table}[H]
			\vspace{-3mm}
			\centering
			\def\arraystretch{1.5}
			\tiny
			\begin{tabular}{|l|l|l|l|l|}
				\hline
				& Percision           & Recall            & F1              & Accuracy             \\ \hline
				Unison  & 84.7 & 83.4          & 84.1 & 85          \\ \hline
				SVM\_Behavior     & 80.3          & 82.2          & 81.2           & 80 \\ \hline
				EmoDNN &\textbf{86}  & \textbf{85.6} & \textbf{85.8}            & \textbf{85.8} \\ \hline
			\end{tabular}
			\vspace{-2mm}
			\caption{Compare Baselines in MultiClass Dataset}
			\label{tab:MultiClass_compare}
			\vspace{-5mm}
		\end{table}
		\begin{figure}[H]
			\vspace{-3mm}
			\centering
			\includegraphics[width=0.40\linewidth]{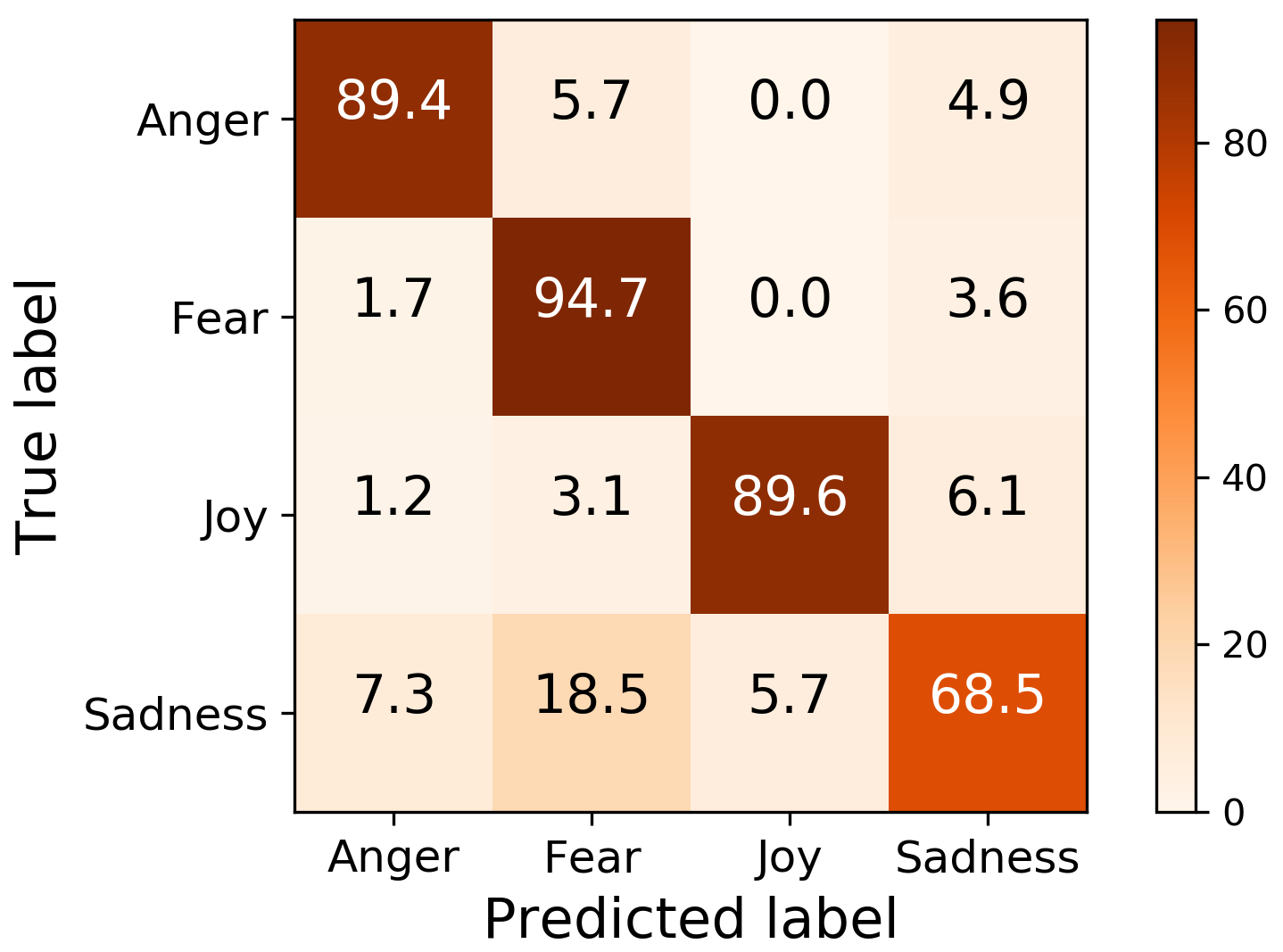}
			\vspace{-3mm}
			\caption{Confusion matrix for emotion prediction}
			\label{fig:Confusion_Matrix}
			\vspace{-4mm}
		\end{figure}
		\par \noindent
		Fig. \ref{fig:Confusion_Matrix} illustrates the prediction confusion matrix for our model, where the weights show the percentage of the correctly predicted samples. EmoDNN has successfully recognized fear in 94.7\% of the labeled tweets. Where the highest prediction performance is for Fear and Sadness is the least, 68.5\% of the correct labels. Evidently, the sadness is mostly misclassified as fear in more than 18\% of the cases where the least incorrect labels for Joy is 5.7\%. We note that it is almost impossible to predict Fear or Anger by the joy emotion, 0.0\%.
		\vspace{-5mm}
		\subsection{Efficiency}
		\vspace{-1mm}
		\subsubsection{Computational complexity analysis}
		\vspace{-1mm}
		\label{Computational-Complexity-Analysis}
		Our proposed framework is useful in many real-time applications. Emotion recognition from social contents can better explain the opinion of a community about a product. Also, the recommendation systems can benefit from emotion weights to improve final suggestions. Many applications need to process millions of brief contents that make the efficiency of the emotion-aware inference systems critical. Hence, we examine the time requirement of our framework.\\
		Our proposed framework comprises offline and online components, where the latter is more complex than the former. In retrospect, we can calculate the complexity for the offline section by aggregating the times of including components. Given $m$ as the number for training samples, the complexity for the SVR-based module to infer the cognitive factors will be $O(m^3)$. The complexity pertaining to two other components, cognitive categorization, and multi-channel feature extraction can be respectively computed as $O(m)$ and $O(m.\Gamma)$, with $\Gamma$ denoting the number of words in each training samples. Let $l$ and $k$ be the respective index and the number of convolutional layers. In that case, the expected time for our network to run each category will thus yield in $O((\sum_{l=1}^k n_{l-1}. |w_l|^2.n_l.Q_l^2).m.e)$ \cite{He2015}. Where $n_l$ and $n_{l-1}$ will respectively represent the number of filters and input channels for the $l^{th}$ layer, $|w_l|$ can signify the filter length, $Q_l$, the output feature spatial size, and $e$, the number of epochs. Correspondingly, the time complexity for the cognitive ensemble can be aggregated by all cognitive categories, verbalized as $O(2q .(\sum_{l=1}^k n_{l-1}. |w_l|^2.n_l.Q_l^2).m.e)$. Also, We designate $q$ with 5 as the number of cognitive cues that as a small constant can be dismissed in time function. Where the time complexity applies to both training and testing times, though with a different scale, the weights can differ in the attention feature vector and the canonical layers of the emotion recognition network. Suppose $\alpha$, $\gamma$, and $\beta$ are constant multipliers. Hence, Eq. \ref{eq:times_complexity} can formalize the overall time complexity of our framework.\\
		However, we further need to efficiently infer millions of brief messages in the online phase. Depending on the network structure, since the time complexity of the online section is polynomial, our model can satisfy the efficacy requirements for real-time processing. The time complexity of the online section to exploit emotions from a single short text is represented by $O(\sum_{l=1}^k n_{l-1}. |w_l|^2.n_l.Q_l^2)$.
		\vspace{-3mm}
		\begin{equation} 
			\small
			\label{eq:times_complexity}
			\begin{gathered}
				\alpha m^3 +m.\Gamma +\gamma m+ \beta(\sum_{l=1}^k n_{l-1}. |w_l|^2.n_l.Q_l^2).m.e \simeq \\[-5pt]
				\alpha m^3+ \beta(\sum_{l=1}^k n_{l-1}. |w_l|^2.n_l.Q_l^2).m.e
			\end{gathered}
			\vspace{-5mm}
		\end{equation}
		\vspace{-6mm}	
		\section{Conclusion}
		\label{conclusion}
		\vspace{-1mm}
		Our proposed unified framework in this paper leverages individual cognitive cues to recognize emotions from short text contents. Most previous efforts on emotion recognition disregard user-specific characteristics. To fill the gap, we firstly categorize short texts according to the cognitive cues. Subsequently, we then utilize the emotion lexicons alongside embedding models to obtain the emotion-aware short text vectors. Consequently, we learn corresponding base classifiers and employ a novel ensemble learning approach to aggregate the classification outputs. The results from extensive experiments on real-world datasets confirm the superiority of our proposed framework over state-of-the-art rivals in emotion recognition. However, we need to integrate transfer learning to make inner ensemble classifiers better collaborate. Moreover, we will have to empirically study the effect of various distributions on the proposed dropout module. We leave these tasks for future work.		
		\vspace{-3mm}
		\ifCLASSOPTIONcaptionsoff
		\newpage
		\fi
		\vspace{-2mm}

		\bibliographystyle{IEEEtran}
		\bibliography{IEEEexample}
\vspace{-13mm}
\begin{IEEEbiography}[{\includegraphics[width=1.0in,height=1.25in,clip,keepaspectratio]{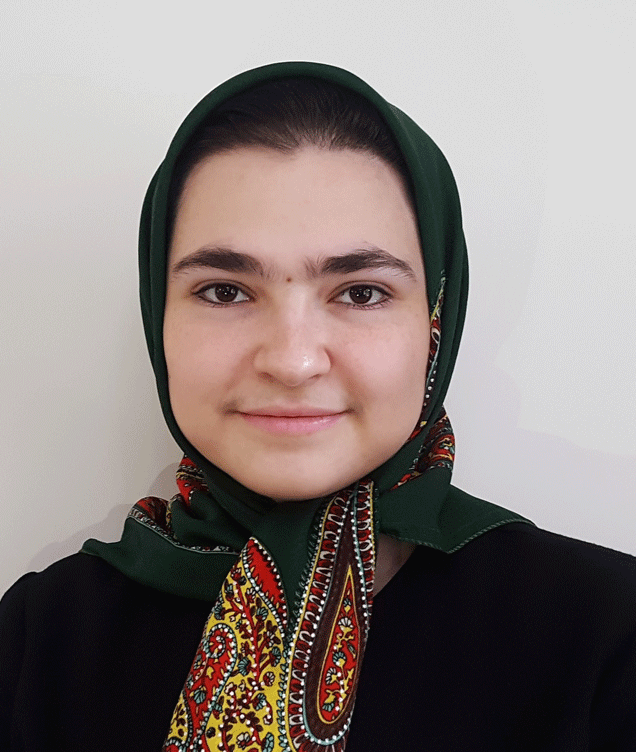}}]{Sara Kamran}
	%\begin{IEEEbiographynophoto}{Sara Kamran}
	is a researcher in the Computational Cognitive laboratory of Iran University of Science and Technology (IUST). She received M.Sc. from IUST and B.S. from the Urmia University of Technology, Iran, in software engineering. Her research interests include affective and cognitive computing, Human-Computer-Interaction, NLP, machine learning, and data analysis.
\end{IEEEbiography}
\vspace{-15mm}
\begin{IEEEbiography}[{\includegraphics[width=1.0in,height=1.25in,clip,keepaspectratio]{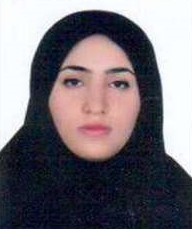}}]{Raziyeh Zall}
	%\begin{IEEEbiographynophoto}{Raziyeh Zall}	
	received the B.S degree from University of Shahid Beheshti of Iran, and M.Sc degree from the Alzahra university of Iran. She is currently working toward the PHD degree in computational cognitive models laboratory at Iran University of Science and Technology. Her research interests include affective and cognitive computing, NLP, and Multi view learning.
\end{IEEEbiography}
\vspace{-15mm}
\begin{IEEEbiography}[{\includegraphics[width=1.0in,height=1.25in,clip,keepaspectratio]{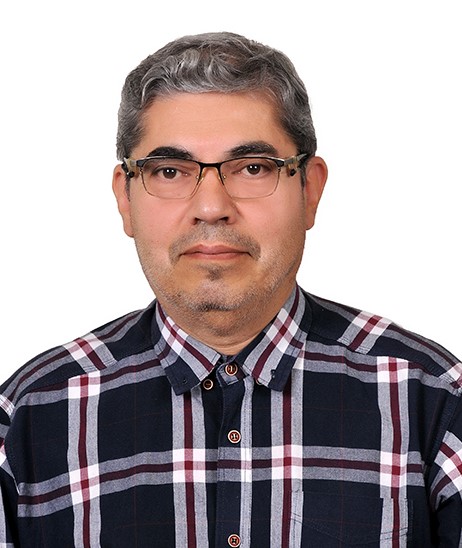}}]{Mohammad Reza Kangavari}
	%\begin{IEEEbiographynophoto}{Mohammad Reza Kangavari}	
	received B.Sc. in computer science from the Sharif University of Technology, M.Sc. from Salford, and Ph.D. from the University of Manchester. He is an associate professor at the Iran University of Science and Technology. His research interests include Intelligent Systems, Human-Computer-Interaction, Cognitive Computing, Machine Learning, and Sensor Networks.
\end{IEEEbiography}
\vspace{-14mm}
\begin{IEEEbiography}[{\includegraphics[width=1.0in,height=1.25in,clip,keepaspectratio]{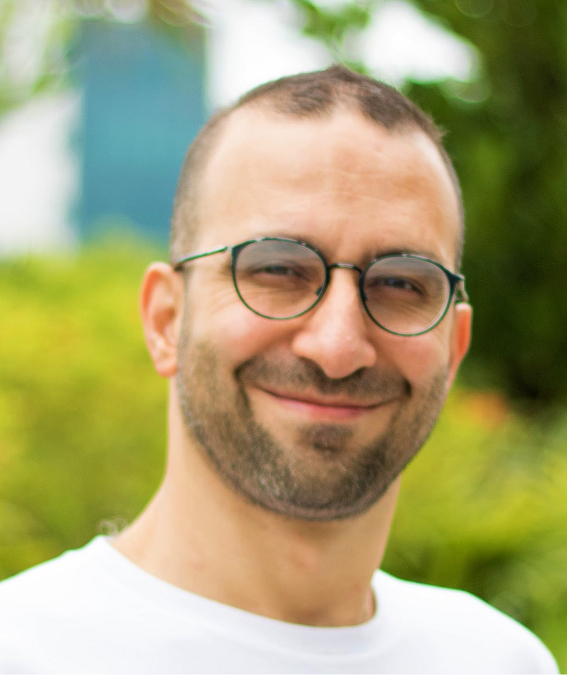}}]{Saeid Hosseini}
	%\begin{IEEEbiographynophoto}{Saeid Hosseini}
	currently works as an assistant professor at Sohar University. He won the Australian Postgraduate Award and received Ph.D. degree in Computer Science from the University of Queensland, Australia, in 2017. He has also completed two post docs in Singapore and Iran. His research interests include spatiotemporal database, dynamical processes, data and graph mining, big data analytics, recommendation systems, and machine learning.
\end{IEEEbiography}
\vspace{-16mm}
\begin{IEEEbiography}[{\includegraphics[width=1.0in,height=1.25in,clip,keepaspectratio]{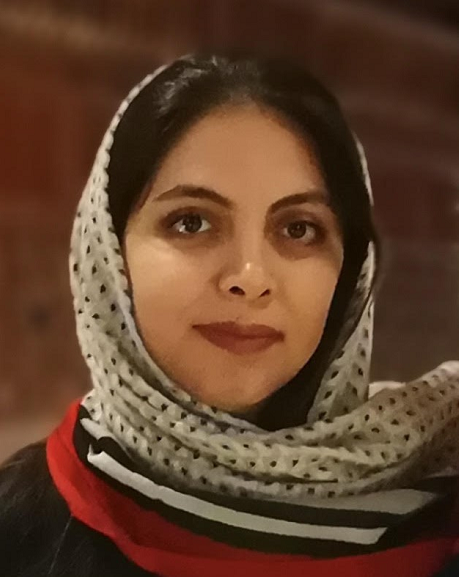}}]{Sana Rahmani}
	%\begin{IEEEbiographynophoto}{Sana Rahmani}	
	is a researcher in the Computational Cognitive laboratory of Iran University of Science and Technology (IUST). She received M.Sc. from IUST and B.S. from the University of Kurdistan, Iran, in software engineering. Her research interests include multimodal affective analysis, Human-Computer-Interaction, machine learning, and data analysis.
\end{IEEEbiography}
\vspace{-14mm}
\begin{IEEEbiography}[{\includegraphics[width=1in,height=1.25in,clip,keepaspectratio]{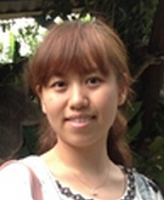}}]{Wen Hua}
	%\begin{IEEEbiographynophoto}{Wen Hua}
	
	currently works as a Lecturer at the University of Queensland. She received her doctoral and bachelor degrees in Computer Science from Renmin University. Her current research interests include natural language processing, information extraction and retrieval, text mining, social media analysis, and spatiotemporal data analytics. She has published articles in reputed venues including SIGMOD, TKDE, VLDBJ.
\end{IEEEbiography}
%\begin{IEEEbiographynophoto}{----}
%	Biography text here.

%\newpage

%\begin{IEEEbiographynophoto}{----}
%	Biography text here.


%\vfill


\end{document}